\documentclass[11pt]{article}

\usepackage[final]{acl}

\usepackage{times}
\usepackage{latexsym}

\usepackage[T1]{fontenc}

\usepackage[utf8]{inputenc}

\usepackage{microtype}

\usepackage{inconsolata}

\usepackage{graphicx}
\usepackage{hyperref}       
\usepackage{url}            
\usepackage{booktabs}       
\usepackage{amsfonts}       
\usepackage{nicefrac}       
\usepackage{microtype}      
\usepackage[table]{xcolor}
\usepackage{xcolor}         
\usepackage{graphicx}
\usepackage{natbib}
\usepackage{multicol}
\usepackage{multirow}
\usepackage{color}
\usepackage{adjustbox}
\usepackage[most]{tcolorbox}
\usepackage{pifont}
\usepackage{alltt}
\usepackage{tabularx}
\usepackage{multirow}
\usepackage{longtable}
\usepackage{placeins} 
\usepackage{fvextra}
\usepackage{afterpage}
\usepackage{cuted}
\usepackage{tcolorbox}

\tcbset{
  aibox/.style={
    width=\linewidth,   
    top=10pt,
    colback=white,
    colframe=black,
    colbacktitle=black,
    enhanced,
    center,
    attach boxed title to top left={yshift=-0.1in,xshift=0.15in},
    boxed title style={boxrule=0pt,colframe=white,},
  }
}

\newtcolorbox{AIbox}[2][]{aibox,title=#2,#1}

\newcommand{\takeawaybox}[1]{%
  \begin{tcolorbox}[
    colback=gray!20!white,    
    boxrule=0pt,
    arc=0mm,
    left=5pt,
    right=5pt,
    top=2pt,
    bottom=2pt,
    fontupper={\fontsize{10pt}{11.75pt}\selectfont}
  ]
    \textbf{Takeaway:} #1
  \end{tcolorbox}%
  \vspace*{-0.1cm}
}

\newcommand{\promptbox}[1]{%
  \begin{tcolorbox}[
    colback=gray!20!white,    
    boxrule=0pt,
    arc=0mm,
    left=5pt,
    right=5pt,
    top=2pt,
    bottom=2pt,
    fontupper={\fontsize{10pt}{11.75pt}\selectfont}
  ]
    \textbf{Prompt:} #1
  \end{tcolorbox}%
  \vspace*{-0.1cm}
}

\usepackage[skip=2pt]{caption}
\usepackage{enumitem}
\setlist[itemize]{noitemsep, topsep=0pt, partopsep=0pt}
\usepackage{xspace}
\newcommand{\benchmark}{{RiskCueBench}\xspace}
\usepackage{cleveref}

\usepackage{mathtools}

\usepackage{float}

\title{RiskCueBench: Evaluating Vision Language Models for Anticipatory Risk Reasoning in Videos}

\author{
  \textbf{Sha Luo}\textsuperscript{1,*}\quad
  \textbf{Yogesh Prabhu}\textsuperscript{2,*}\quad
  \textbf{Timothy Ossowski}\textsuperscript{1,*}\quad
  \textbf{Kaiping Chen}\textsuperscript{1,\textdagger}\quad
  \textbf{Junjie Hu}\textsuperscript{1,\textdagger} \\
  \textsuperscript{1}University of Wisconsin--Madison, Madison, WI, USA \\
  \textsuperscript{2}University of California San Diego, San Diego, CA, USA \\
  \texttt{sluo83@wisc.edu, ossowski@wisc.edu, yprabhu@ucsd.edu,}\\ \texttt{kchen67@wisc.edu, jhu@cs.wisc.edu} 
}

\begin{document}
\maketitle

\begin{abstract}

Anticipating risky events before accidents unfold is essential for safety-critical AI systems and increasingly relevant in video-centric social media. However, existing video risk assessment benchmarks largely evaluate post-hoc understanding, assuming access to complete videos that include the incident itself. We introduce \benchmark, a benchmark spanning four risk domains that evaluates whether vision-language models (VLMs) can anticipate emerging safety concerns from incomplete visual cues. Each video is annotated with a risk signal clip, defined as the earliest segment containing sufficient cues for humans to infer elevated risk prior to incident onset. To assess not only prediction accuracy but also reasoning quality, we adopt a Question-Reasoning-Answer (QRA) framework that evaluates grounded reasoning, temporal sensitivity, and self-correction behavior. Experiments across 16 state-of-the-art VLMs reveal a substantial gap between current systems (best F1 0.69) and human performance (F1 0.98), with particularly pronounced failures on socially embedded risks such as protests and arson. These results show persistent limitations in anticipatory video reasoning and establish \benchmark as a testbed for safety-oriented video understanding.
\end{abstract}
\begin{table*}[!ht]
\centering
\resizebox{\textwidth}{!}{%
\begin{tabular}{lccccc}
\toprule
\textbf{Dataset} & \textbf{Temporal Reasoning} & \textbf{Object Grounding} & 
\textbf{Reasoning Metrics} & \textbf{Event Forecasting} & \textbf{Risk Events}\\
 & \S\ref{sec:temporal_reasoning} & \S\ref{sec:reasoning_grounding} & \S\ref{sec:overthinking} & \S\ref{sec:forecasting} & \S\ref{sec:annotation} \\
\midrule
TimeLogic & \ding{51} & \ding{51} & \ding{55} & \ding{55} & \ding{55}\\
ReXTime & \ding{51} & \ding{51} & \ding{55} & \ding{55} & \ding{55}\\
MotionBench  & \ding{51} & \ding{55} & \ding{55} & \ding{55} &\ding{55}\\
MVBench & \ding{51} & \ding{51} & \ding{55} & \ding{55} & \ding{55} \\
VLM4D & \ding{51} & \ding{55} &  \ding{55} & \ding{55} & \ding{55} \\
RTV-Bench & \ding{51} &  \ding{51}  & \ding{55} & \ding{51} & \ding{55} \\
FutureBench & \ding{51} &  \ding{55}  & \ding{55} & \ding{51} & \ding{55} \\
RiskBench & \ding{51} &  \ding{51}  & \ding{55} & \ding{51} & \ding{51} \\
\midrule
\textbf{Ours} & \ding{51} & \ding{51} & \ding{51} &\ding{51} & \ding{51} \\
\bottomrule
\end{tabular}
}
\caption{Overview of benchmarks for spatio-temporal reasoning in vision–language models. Unlike prior work, our analysis centers on risk events, evaluating models’ ability to forecast under uncertainty, while also examining reasoning traces and object grounding.}
\label{tab:video-benchmarks}
\vspace{-0.5em}
\end{table*}

\section{Introduction}







\begin{figure}[t]
    \centering
    \includegraphics[width=0.8\linewidth]{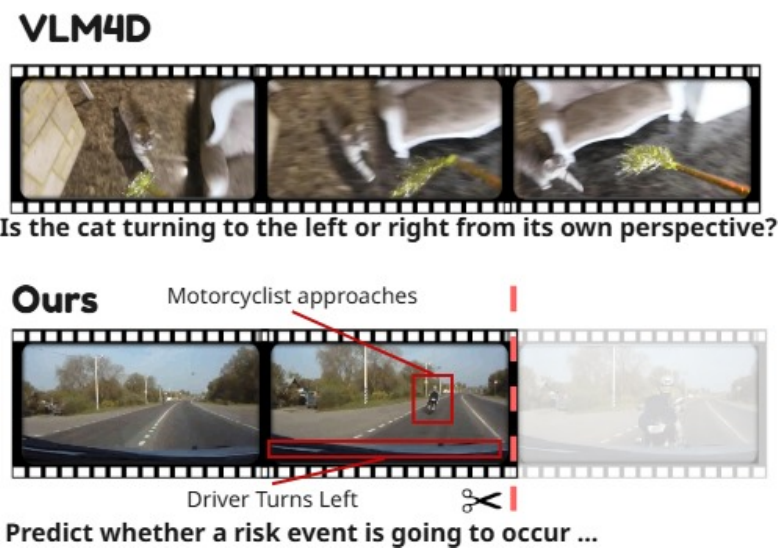}
    \caption{Many existing benchmarks ask about events that occur during the video. Our \benchmark focuses on predicting future risky events (details in \S\ref{sec:benchmark}).}
    \label{fig:benchmark_comparison}
    \vspace{-4mm}
\end{figure}

Advanced vision-language models (VLMs) have
surfaced with remarkable abilities to comprehend
complex spatial relationships, temporal sequences,
and visual narratives. However, the specific needs
of safety-critical applications, particularly situa-
tional risk assessment and prediction, still differ
significantly from those of general video under-
standing benchmarks. As shown in Figure \ref{fig:benchmark_comparison}, most
existing benchmarks emphasize post hoc under-
standing, where models analyze or describe events
after they have fully unfolded, such as answering
questions or generating captions given complete
visual context \cite{caba2015activitynet, Zhou_2025_ICCV, Hong_2025_CVPR}. 

Predictive reasoning requires models to anticipate future events from ambiguous early signals. While current VLMs perform well on descriptive tasks, their capacity for such anticipatory reasoning remains largely unexplored. Recent benchmarks targeting next event prediction typically frame evaluation as multiple choice question answering, where models select from predefined outcomes rather than reason freely about future risks \cite{wu2024star, wang2025fostering, xun2025rtv, cheng2025video}, which lack the open-ended nature of real-world risk. Other work focuses narrowly on traffic accidents via synthetic data or non-reasoning metrics \cite{kung2024riskbench, fatima2021global, bao2020uncertainty, hussain2024bi}, leaving broader risk types such as crowd dynamics, malicious intent in arson, sudden kinematic hazards in traffic, and complex navigational failures in cruise collisions, largely unaddressed. Covering these domains is essential to study whether anticipatory reasoning generalizes across physical kinematics and social behaviors.

To bridge this gap, we propose \benchmark, a benchmark evaluating whether VLMs can anticipate emerging safety concerns from early video signals (Table~\ref{tab:video-benchmarks}). Using a Question-Reasoning-Answer (QRA) framework, we explicitly capture model decision-making. Our analysis reveals significant limitations in state-of-the-art VLMs' predictive abilities. Our contributions include:

\begin{itemize}[leftmargin=13pt,noitemsep]
   \item We introduce \benchmark, a VLM benchmark of 1648 videos across four domains (protest, car crash, arson, cruise collision), requiring inference of danger from subtle pre-incident cues.
    \item We develop a workflow that uses model disagreement and LLM filtering to automatically curate challenging high-utility video examples (\S\ref{sec:benchmark}).
    \item We propose three diagnostic metrics: Reasoning Grounding Accuracy (RGA), Temporal Reasoning Difference (TRD), and Self-Correction Degradation (SCD), to quantify VLM reasoning behavior (\S\ref{sec:analysis}).
    \item We show that state-of-the-art VLMs substantially underperform on socially-embedded risks (protests/arson) compared to kinematic ones (car/cruise), revealing key anticipatory failures.
\end{itemize}

\begin{figure*}[t]
    \centering
    \includegraphics[width=\textwidth]{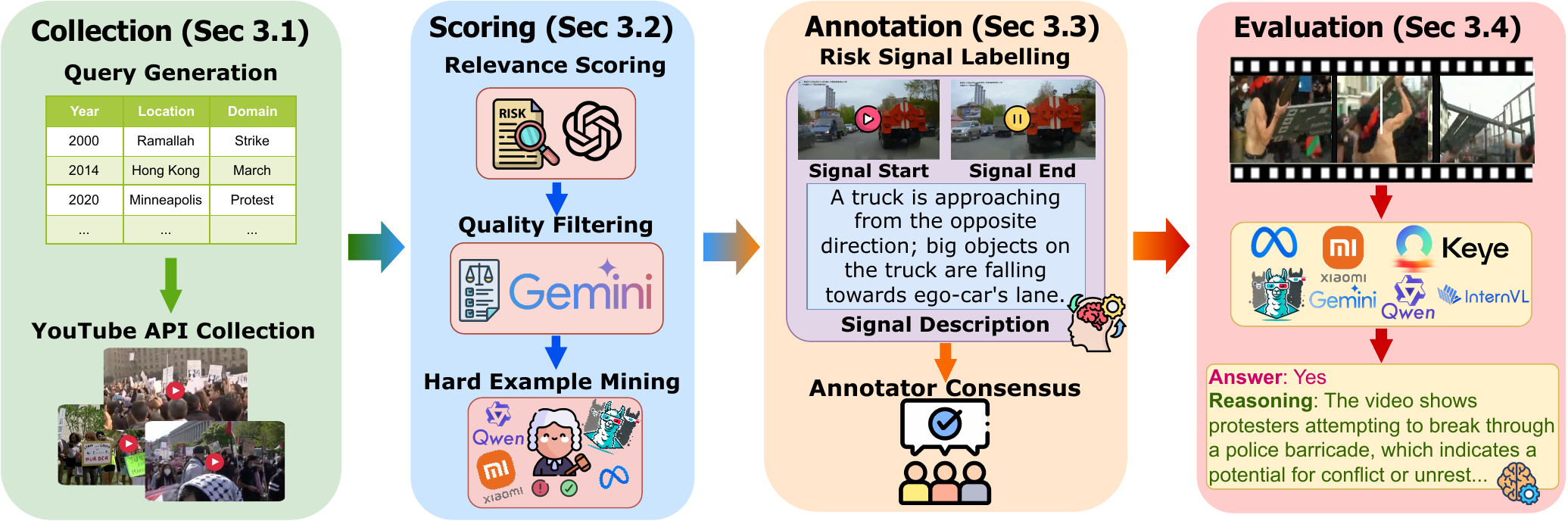}
    \caption{Overview of our pipeline to curate risk signal clips from real-world incidents and evaluate VLM reasoning. 
    \textbf{Collection: } We collect a large candidate set of YouTube videos using domain-relevant keywords. \textbf{Scoring:} The collected videos are filtered to only retain those with potential risk and high difficulty. \textbf{Annotation: } Human annotators label the filtered videos to identify the risk signal clip. \textbf{Evaluation: } Popular VLMs are presented with only the risk signal clip, and their output reasoning traces are evaluated.}
    \label{fig:pipeline}
\end{figure*}

\section{Related Work}

\paragraph{Video Language Models and Reasoning Systems}
Recent advances in VLM encoding for temporal reasoning which combines language modeling with video encoding mechanisms for intricate temporal reasoning is represented by VideoLLaMA 3 \cite{zhang2025videollama}, Video LLaVA \cite{lin2023video}, Qwen-VL \cite{bai2025qwen2}, and Apollo \cite{zohar2025apollo}. With the introduction of reinforcement learning techniques for video comprehension by MiMo-RL \cite{coreteam2025mimovltechnicalreport}, the resolution of long-form video analysis problems by LongVILA-R1 \cite{chen2025scaling}, and the incorporation of explicit reasoning mechanisms that mimic human cognitive processes by GLM-4.1V-Thinking \cite{hong2025glm}, specialized reasoning systems have further advanced this field. 

\paragraph{Spatio-Temporal Reasoning}
Recent work has introduced numerous benchmarks aimed at evaluating the spatio temporal reasoning capabilities of vision language models in video understanding (Table \ref{tab:video-benchmarks}). These efforts probe diverse aspects of temporal and spatial cognition, including temporal logic and event ordering, fine grained motion understanding, and dynamic scene interpretation in domains such as egocentric video and autonomous driving. Benchmarks such as TimeLogic \cite{swetha2025timelogic} and ReXTime \cite{NEURIPS2024_32683193} focus on logical and causal reasoning over event sequences, while MotionBench \cite{Hong_2025_CVPR} and related diagnostic tasks evaluate sensitivity to motion dynamics and temporal direction.  MVBench \cite{li2024mvbench}, VLM4D \cite{Zhou_2025_ICCV}, and STSBench \cite{fruhwirth2025stsbench}, assess joint spatial and temporal reasoning across diverse video scenarios, yet many tasks can still be solved using static cues or retrospective access to the full video.

Crucially, existing benchmarks rely on multiple-choice accuracy, offering limited insight into how models arrive at predictions, they do not assess the faithfulness or grounding of model reasoning. In contrast, our benchmark employs dedicated metrics to assess if risk predictions are supported by temporally coherent and visually grounded explanations. This enables a fine-grained analysis of spatio-temporal reasoning, exposing limitations obscured by traditional accuracy metrics.
\section{Benchmark Construction}
\label{sec:benchmark}


We propose a reproducible, domain-agnostic framework for constructing risk-centric video benchmarks, emphasized by temporally grounded signals. The four-stage process includes: (1) large-scale collection via structured queries (\S\ref{sec:data-collection}); (2) multi-stage filtering for risk relevance (\S\ref{sec:data-filtering}); (3) fine-grained human annotation using a temporal protocol (\S\ref{sec:annotation}); and (4) a set of evaluation metrics for video reasoning, as shown in Figure \ref{fig:pipeline}. While instantiated across four diverse domains (protests, traffic incidents, arson, and cruise collisions), the framework remains easily extensible to other safety-critical scenarios like natural disasters, ensuring both diverse risk coverage and detailed temporal depth.

\subsection{Data Collection}
\label{sec:data-collection}
We first curate diverse risk-relevant text queries for the YouTube Data API and collect an initial pool of candidate videos and their metadata.
\paragraph{Query Curation.} We curate video footage across four domains: protests, traffic incidents, arson, and cruise collisions. To account for differences in how these incidents are visually documented, we implement a dual-stream curation strategy. For macro-events anchored in specific sociopolitical contexts (e.g., protests), we construct queries using GPT-4V generated year–location pairs, followed by manual Google Search verification. For localized micro-anomalies (arson, cruise collisions), use intersectional queries combining action, actor, and target terms to filter out abstracted news. We convert each structured event into a text query using a template in Appendix \ref{app:query_construction}.  This process maximizes coverage while reducing bias toward specific regions or languages, ensuring the dataset reflects diverse sociopolitical and environmental conditions.


\paragraph{Video Collection.} Using these queries, we retrieve videos and their metadata (e.g., title, description, upload date) through the YouTube Data API. This ensured replicable sampling of publicly available video, rather than relying on proprietary or manually sourced data. Metadata enables traceability and allows filtering or stratification by contextual variables such as time and location. 

\subsection{Dataset Filtering}
\label{sec:data-filtering}
We apply a sequence of automated filtering steps to retain videos that depict active risk events and meet minimum visual quality standards.  We use LLMs as high-throughput filtering assistants to manage raw data scale. While these models facilitate candidate selection, the final dataset is established exclusively on the manual protocol in Section 3.3. By keeping human-annotated ground truth as the final authority, we ensure benchmark metrics remain free from systematic LLM bias. 

\paragraph{Text-based Relevance Filtering.} Not all retrieved videos are useful for analyzing real-time risk dynamics; many contain news commentary or post-event summaries. To identify videos that plausibly contain \textit{in-action} risk events, we use an LLM (GPT-4o) to evaluate each video's title and description pair. The model is prompted to assess the likelihood that the video depicts an ongoing risk event rather than commentary, lectures, or summaries:

\promptbox{ On a scale from 1 to 10, how likely is this video to contain [risk event] video in action but not lecture/tutorial/slides, etc.? Answer with a single integer from 1 (very unlikely) to 10 (very likely). }

\noindent Videos scoring below a predefined threshold of 9 out of 10 are filtered out. 

\paragraph{Fine-Grained Visual Quality Filtering.} Even among relevant videos, production artifacts such as subtitles, overlays, or heavy editing can introduce spurious shortcuts for visual models. 
To ensure models perform reasoning on risk-relevant visual content rather than these shortcuts, we use Gemini 2.5 Flash Lite to evaluate videos along 12 visual quality dimensions as in Appendix \ref{tab:quality_scoring} (i.e., logo, location, time/date, reporter presence, SNS overlay, image quality, temporal continuity, consequence text, title/banner, subtitle, camera perspective).
Videos that fail to obtain Score 2 in at least 10 out of the 12 quality criteria are removed, ensuring that the benchmark prioritizes authentic, unedited videos and emphasizes visual perception and reasoning under realistic conditions.

\paragraph{Hard Example Mining.} To explicitly include challenging and ambiguous cases, we perform hard example mining using a committee of five VLMs (Gemini 2.5, Qwen-7B, MiMo-RL, MiMo-SFT, InternVL) across all four domains. Given a standardized visual-only prompt (Appendix \ref{12critera_prompt}), we define a \textit{hard example} as a video if more than three out
of five models disagree with each other. These high-disagreement cases represent ambiguous or complex scenarios which challenge both human and AI perception, strengthening the dataset’s ability to test model resilience in borderline conditions.

\subsection{Annotation}
\label{sec:annotation}

The final dataset is annotated utilizing crowdsourced workers via Amazon Mechanical Turk (MTurk) and trained internal annotators using a structured protocol designed to capture the temporal and visual progression of risk (the full protocol, instruction materials and interface documentation are included in Appendix \ref{app:annotation}).

\paragraph{Annotation Protocol.} Each video is annotated along eight dimensions that jointly consider temporal boundaries, visual cues, and semantic content (annotation example see Appendix \ref{app:annotation_examples})

\begin{enumerate}[leftmargin=13pt,noitemsep]
    \item \textbf{Risk Signal Start.} Timestamp of the first observable visual signal indicating potential risk (or no risk). 
    \item \textbf{Risk Signal End.} Timestamp when clear risk (or no-risk) indicators cease. 
    \item \textbf{Risk Visual Indicator.} Initial visual cues used to judge risk/no risk. 
    \item \textbf{Risk Signal Description.} Full narrative description of the signals using temporal markers (\textit{first}, \textit{then}, \textit{afterwards}). 
    \item \textbf{Accident Start Frame.} Timestamp of the first observable moment of incident. 
    \item \textbf{Accident End Frame.} Timestamp when the incident or scene fully concludes. 
    \item \textbf{Accident Description.} One or more sentences summarizing the incident. 
    \item \textbf{Risk Label.} Binary assignment of ``yes'' (clear risk observed) or ``no'' (e.g., peaceful protest). 
    
\end{enumerate}

\noindent To ensure data quality and objectivity across the MTurk workforce, we conducted an Inter-Annotator Agreement validation study. A randomly sampled subset of the videos was independently labeled by internal expert annotators and cross-referenced with the crowdsourced annotations. For categorical risk labels, we achieved $100\%$ agreement. For temporal boundaries, we observed a Mean Absolute Deviation of less than $4$ frames. This process ensures the dataset retains high-granularity temporal and semantic information regarding risk onset, escalation, and resolution. 

\paragraph{Instantiation.} Using our pipeline, we collect video data from four risk domains: protests, traffic incidents, arson, and cruise collisions. We report statistics in Table 2. Depending on the domain and data source, certain steps in the pipeline may be omitted. For example, a 50-video subset of the arson domain were sourced from existing, preprocessed datasets \cite{sultani2018real}, we omitted preliminary steps such as YouTube API collection and initial visual quality filtering for it.\\


\begin{table}[!h]
\centering

\small

\resizebox{\columnwidth}{!}{%
\begin{tabular}{lcc}
\toprule
Video Type & Average Signal Length (s) & \# Videos \\
\midrule
Car Crash (Normal) & 3.49 $\pm$ 0.58 & 250 \\
Car Crash (High Risk) & 1.38 $\pm$ 0.77 & 252 \\
Protest (Normal) & 17.77 $\pm$ 14.81 & 267 \\
Protest (High Risk) & 5.99 $\pm$ 7.07 & 217 \\
Arson (Normal) & 15.83 $\pm$ 26.24 & 178 \\
Arson (High Risk) & 17.33 $\pm$ 26.29 & 231 \\
Cruise Collision (Normal) & 44.68$\pm$29.27  & 134 \\
Cruise Collision (High Risk) & 42.51$\pm$34.46 & 119 \\
\midrule 
Total & & 1648 \\
\bottomrule
\end{tabular}}

\caption{Risk signal length for \benchmark videos.}
\label{tab:dataset_statistics}

\end{table}

\noindent Risk signal lengths vary by domain, e.g, Cruise Collision events average 43.6 seconds, reflecting gradual escalation, while Car Crash signals average just 2.4 seconds due to their abrupt nature (see Appendix \ref{fig:combined_risk_length}). These statistics define the "decision windows" available for VLMs to predict risk. Because risk and non-risk clips also differ in duration, we conduct an additional length-matched analysis to assess whether clip duration drives model performance. Results remain highly stable after matching, suggesting that the observed performance limitations are not primarily explained by duration imbalance (Appendix~\ref{app:length_matching}).

\subsection{Evaluation}
\label{sec:evaluation}

\paragraph{Risk Prediction (F1).} We first evaluate the binary risk classification of each model using the standard F1 score.


\paragraph{Reasoning Grounding Accuracy (RGA).}
To quantify how well model reasoning is anchored in relevant visual evidence, we calculate the semantic alignment between model-extracted reasoning items ($\hat{\mathcal{O}}_i = \{\hat{o}_{i1}, \ldots, \hat{o}_{ij}\}$) and human-annotated risk indicators ($\mathcal{O}_i = \{o_{i1}, \ldots, o_{ik}\}$) using SentenceTransformer embeddings (all-MiniLM-L6-v2). Here, $i$ indexes the video sample, $j$ indexes predicted decision items, and $k$ indexes ground-truth visual indicators. For a decision item $\hat{o}_{ij}$, we compute a \textbf{relatedness score} by its maximum cosine similarity against all ground-truth indicators:
\begin{equation}
    s_{ij} = \max_{k \in \{1,\ldots,m\}} \cos(\mathbf{e}_{\hat{o}_{ij}}, \mathbf{e}_{o_{ik}})
    \label{rga_eq:1}
\end{equation}
where $\mathbf{e}_{\hat{o}_{ij}}$ and $\mathbf{e}_{o_{ik}}$ denote the embedding vectors for decision item $\hat{o}_{ij}$ and visual indicator $o_{ik}$, respectively. Because $s_{ij}$ is continuous, we use a threshold $\tau$ to determine whether a decision item is treated as semantically grounded:
\begin{equation}
    g_{ij} = \mathbb{I}\left[s_{ij} \geq \tau\right]
\end{equation}
we select $\tau$ through an F1-maximization sweep using the association between average semantic similarity scores and binary prediction correctness. We use this procedure as a practical calibration strategy for separating higher- and lower-relatedness reasoning items. We note that $\tau$ is a task-specific operating threshold rather than a unique threshold for semantic grounding, and we report additional robustness metrics in Appendix \ref{app:additional_metrics}. Finally, RGA is the average proportion of correctly grounded decision items across all sampled test samples:
\begin{equation}
    \text{RGA} = \frac{1}{|\mathcal{V}|} \sum_{v_i \in \mathcal{V}} \frac{1}{|\hat{\mathcal{O}}_i|} \sum_{j=1}^{|\hat{\mathcal{O}}_i|} g_{ij} \times 100\%
\end{equation}
To assess the reliability of this LLM-assisted grounding evaluation, we additionally conduct a human validation on high- and low-RGA samples. Human judgments show clear separation between the two groups, supporting RGA as a meaningful proxy for visual grounding quality (Appendix~\ref{app:rga_validation}).

\paragraph{Temporal Reasoning Difference (TRD).} To measure model sensitivity to temporal ordering, we augmented each video in 3 different ways and observed the effect on performance:
\begin{itemize}[noitemsep]
    \item \textbf{Frames Shuffled:} All the frames in the video are shuffled into random order.
    \item \textbf{Half-Swapped:} The first half of the video is swapped with the second half.
    \item \textbf{Frames Reversed:} The frames of the video are presented to the model in reverse order.
\end{itemize}

For each augmentation type $a \in \mathcal{A} = \{\text{``shuffled''}, \text{``swapped''}, \text{``reversed''}\}$, we compute the absolute F1 score difference between the original video and its augmented counterpart. The TRD metric is then calculated as the average absolute F1 difference across all augmentation types:
\begin{equation}
    \text{TRD} = \frac{1}{|\mathcal{A}|} \sum_{a \in \mathcal{A}} \left| \text{F1}_{\text{original}} - \text{F1}_{a} \right|
\end{equation}
A higher TRD value indicates sensitivity to temporal ordering, suggesting the model relies on temporal information for risk prediction. Conversely, a TRD near zero implies performance changes little when temporal order is disrupted. Noting that low TRD does not imply that individual frames contain no predictive information. To distinguish temporal-order sensitivity from sparse static-cue reliance, we additionally evaluate first-frame-only and last-frame-only baselines (Appendix~\ref{app:keyframe_analysis}). On average, the first-frame condition underperforms the full-video condition across domains, whereas last-frame performance is generally comparable to the full-video condition. This is expected because the last frame immediately precedes incident onset and therefore contains more explicit pre-incident cues. Since shuffling and reversal preserve such salient frames while disrupting their temporal progression, low TRD together with the key-frame results suggests that models rely substantially on the presence of salient later-stage cues while making more limited use of the temporal ordering through which those cues emerge.


\paragraph{Self-Correction Degradation (SCD).} To quantify the performance difference associated with explicit self-correction behavior, we partition the dataset into two subsets based on whether the judge-extracted confusion count is non-zero: $\mathcal{V}^{+} = \{v_i \mid c_i > 0\}$ (samples with self-correction markers) and $\mathcal{V}^{-} = \{v_i \mid c_i = 0\}$ (samples without). We then compute a weighted F1 gap that accounts for subset size:
\begin{equation}
\text{SCD} = P(\mathcal{V}^{+}) \cdot \text{F1}(\mathcal{V}^{+}) - P(\mathcal{V}^{-}) \cdot \text{F1}(\mathcal{V}^{-})
\end{equation}
where $P(\mathcal{V}^{k}) = \frac{|\mathcal{V}^{k}|}{|\mathcal{V}|}$ is the proportion of samples in subset $k$, and $\text{F1}(\cdot)$ denotes the F1 score over the respective subset. The weighting reflects differences in subset prevalence when comparing self-correction patterns across models. Importantly, SCD is defined by the presence of explicit self-correction markers rather than reasoning length. Our additional analysis shows that responses containing such markers tend to have longer reasoning chains, indicating that reasoning length and example difficulty may covary with self-correction behavior. A negative SCD value indicates lower weighted F1 performance for the marker-present subset, with more negative values reflecting a larger performance gap associated with self-correction markers.


\section{Experiments and Analysis}
\label{sec:analysis}
\subsection{Experimental Setup}
Videos are sampled at 1 frame per second (fps) to balance computational efficiency with temporal
coverage. All models use a standardized prompt
template; for reasoning-enhanced models, we
omit explicit instructions as these capabilities
are integrated into their existing chat templates.

\paragraph{Baseline Models} 
We evaluate 16 SOTA models. \\
\textit{(1) Standard VLMs} perform direct video-to-text generation; include Video LLaVA, VideoLLaMA 3, InternVL 3.5, Qwen-VL, ChatUniVi, Gemini-Flash-3, Apollo, and MiMo-SFT. \\
\textit{(2) Reasoning-Enhanced VLMs} leverage explicit reasoning chains or reinforcement learning-based training to bolster predictive accuracy. These are represented by Qwen-VL Thinking, MiMo-RL, LongVILA-R1, and GLM-4.1V-Thinking.



\subsection{Overall Performance}

\begin{table*}[h!]
    \centering
    \footnotesize
    \resizebox{\textwidth}{!}{
    \begin{tabular}{lcccccccccccccccc}
    \toprule
    \multirow{2}{*}{Model} 
    & \multicolumn{4}{c}{\textbf{Car Crash}} 
    & \multicolumn{4}{c}{\textbf{Protest}} 
    & \multicolumn{4}{c}{\textbf{Arson}}
    & \multicolumn{4}{c}{\textbf{Cruise Collision}} \\ 
    \cmidrule(lr){2-5} \cmidrule(lr){6-9} \cmidrule(lr){10-13} \cmidrule(lr){14-17}
     & F1 & RGA & TRD & SCD 
     & F1 & RGA & TRD & SCD 
     & F1 & RGA & TRD & SCD 
     & F1 & RGA & TRD & SCD \\ 
    \midrule
    \multicolumn{17}{l}{\textit{\textbf{Non-Reasoning Models}}} \\
    Video-LLaVA         & 0.53 & 48.2 & 1.84 & - & 0.44 & 33.5 & 1.87 & - & 0.55 & 52.8 & 1.91 & - & 0.55 & 49.3 & 1.68 & - \\
    VideoLLaMA 3        & 0.59 & 54.1 & 1.65 & - & 0.46 & 39.2 & 2.42 & - & 0.57 & 53.2 & 1.65 & - & 0.56 & 52.0 & 1.61 & - \\
    Qwen3-VL-8B         & 0.64 & 57.3 & 1.48 & - & 0.48 & 42.8 & 2.23 & - & 0.62 & 55.3 & 1.59 & - & 0.63 & 57.7 & 1.72 & - \\
    MiMo-SFT            & 0.58 & 51.4 & 2.12 & - & 0.47 & 38.1 & 1.76 & - & 0.56 & 50.4 & 2.05 & - & 0.59 & 52.5 & 2.08 & - \\
    Apollo-7B           & 0.57 & 53.6 & 1.83 & - & 0.45 & 37.4 & 2.59 & - & 0.58 & 55.6 & 1.82 & - & 0.57 & 52.9 & 1.63 & - \\
    \midrule
    \multicolumn{17}{l}{\textit{\textbf{Reasoning Models}}} \\
    InternVL-3.5        & 0.61 & 64.2 & 2.87 & -0.28 & 0.47 & 47.8 & 2.54 & -0.42 & 0.60 & 63.2 & 2.90 & -0.30 & 0.59 & 62.7 & 2.95 & -0.26 \\
    Qwen3-VL-8B-T       & 0.66 & 68.7 & 2.21 & -0.29 & 0.57 & 56.5 & 2.08 & -0.36 & 0.64 & 66.8 & 2.21 & -0.31 & 0.66 & 69.8 & 2.39 & -0.29 \\
    MiMo-RL             & 0.63 & 61.3 & 3.67 & -0.35 & 0.51 & 49.9 & 3.42 & -0.45 & 0.61 & 60.3 & 3.49 & -0.36 & 0.64 & 61.5 & 3.49 & -0.38 \\
    Keye-VL 1.5         & 0.57 & 63.8 & 3.15 & -0.21 & 0.59 & 52.3 & 2.91 & -0.29 & 0.59 & 52.9 & 2.90 & -0.29 & 0.57 & 57.5 & 2.90 & -0.24 \\
    GLM 4.1 V-T         & 0.58 & 65.1 & 3.39 & -0.26 & 0.56 & 49.1 & 3.18 & -0.34 & 0.58 & 64.1 & 3.31 & -0.27 & 0.56 & 64.6 & 3.51 & -0.23 \\
    \midrule
    \multicolumn{17}{l}{\textit{\textbf{Closed-Source Models}}} \\
    Gemini-Flash-3      & 0.69 & - & - & - & 0.67 & - & - & - & 0.62 & - & - & - & 0.68 & - & - & - \\
    \midrule
    \multicolumn{17}{l}{\textit{\textbf{Baselines}}} \\
    Random Guess        & 0.50 & - & - & - & 0.44 & - & - & - & 0.53 & - & - & - & 0.47 & - & - & - \\
    Human               & 0.98 & - & - & - & 0.97 & - & - & - & 0.98 & - & - & - & 0.96 & - & - & - \\
    \bottomrule
    \end{tabular} }
    \caption{
    Performance statistics of current SOTA models on binary risk prediction across the four domains. F1 denotes positive-class (risk) F1 computed separately within each domain; we do not report an overall or cross-domain average F1. The corresponding always-risk no-skill F1 baselines and additional TN-sensitive metrics including Accuracy, Specificity, and MCC are reported in Appendix 10. Random Guess reports a fixed-seed empirical random baseline; deviations from 0.50 reflect finite-sample variation, class imbalance, and seed noise. Certain metrics remain blank for Gemini-Flash-3 because its internal reasoning traces are not accessible for evaluation.}
    \label{tab:main_results}
\end{table*}

Performance varies significantly across domains (Table.\ref{tab:main_results}). Models achieve F1 scores of 0.27–0.71 on Car Crashes, led by Qwen3-VL-8B (0.7151), followed by InternVL-3.5 (0.6585) and VideoLLaMA 3 (0.6047). Protest scenarios show markedly lower performance (F1: 0.26–0.46), suggesting models struggle with nuanced social and behavioral cues compared to structured vehicular patterns. 

Different models exhibit distinct failure modes. Video LLaVA and ChatUniVi show low precision on protests (high false positives), while MiMo-SFT demonstrates low recall on car crashes (conservative detection). Reasoning-enhanced models do not consistently outperform standard VLMs, suggesting that explicit reasoning mechanisms are not effectively calibrated for early risk signal detection.

Overall, results reveal a critical performance ceiling, particularly for socially-embedded scenarios. F1 scores of 45–55\% on protest footage indicate substantial gaps from human-level performance and fundamental limitations in interpreting early warning signals and predicting safety outcomes from partial information.
\takeawaybox{
    \textbf{Model performance varies sharply by domain.}  While leading vision–language models achieve moderate success on car-crash prediction, they struggle substantially on protest scenarios that require interpreting social and behavioral cues.
}

\subsection{Reasoning Chain Analysis}
To analyze current VLM risk prediction failures, we study two aspects of model behavior: self-correction patterns and reasoning grounding, using Gemini-Pro-3 as a judge.
For each video $v_i \in \mathcal{V}$, we obtain human ground-truth risk annotations $a_i = (d_i, \mathcal{O}_i, l_i)$, where $d_i$ is the human-written risk description, $\mathcal{O}_i = \{o_{i1}, o_{i2}, \ldots, o_{im}\}$ is the set of annotated risk visual indicators, and $l_i$ is the risk label. The judge model parses each model-generated reasoning chain $r_i$ to extract: (1) a confusion count $c_i$, indicating the presence of self-correction markers (e.g., ``wait...'', ``actually...''), and (2) a structured set of predicted decision items $\hat{\mathcal{O}}_i = \{\hat{o}_{i1}, \hat{o}_{i2}, \ldots, \hat{o}_{in}\}$, representing the objects and entities the model references when justifying its individual risk prediction.

\subsubsection{Reasoning Grounding} \label{sec:reasoning_grounding}
\begin{figure}[!h]
    \centering
    \includegraphics[width=\linewidth]{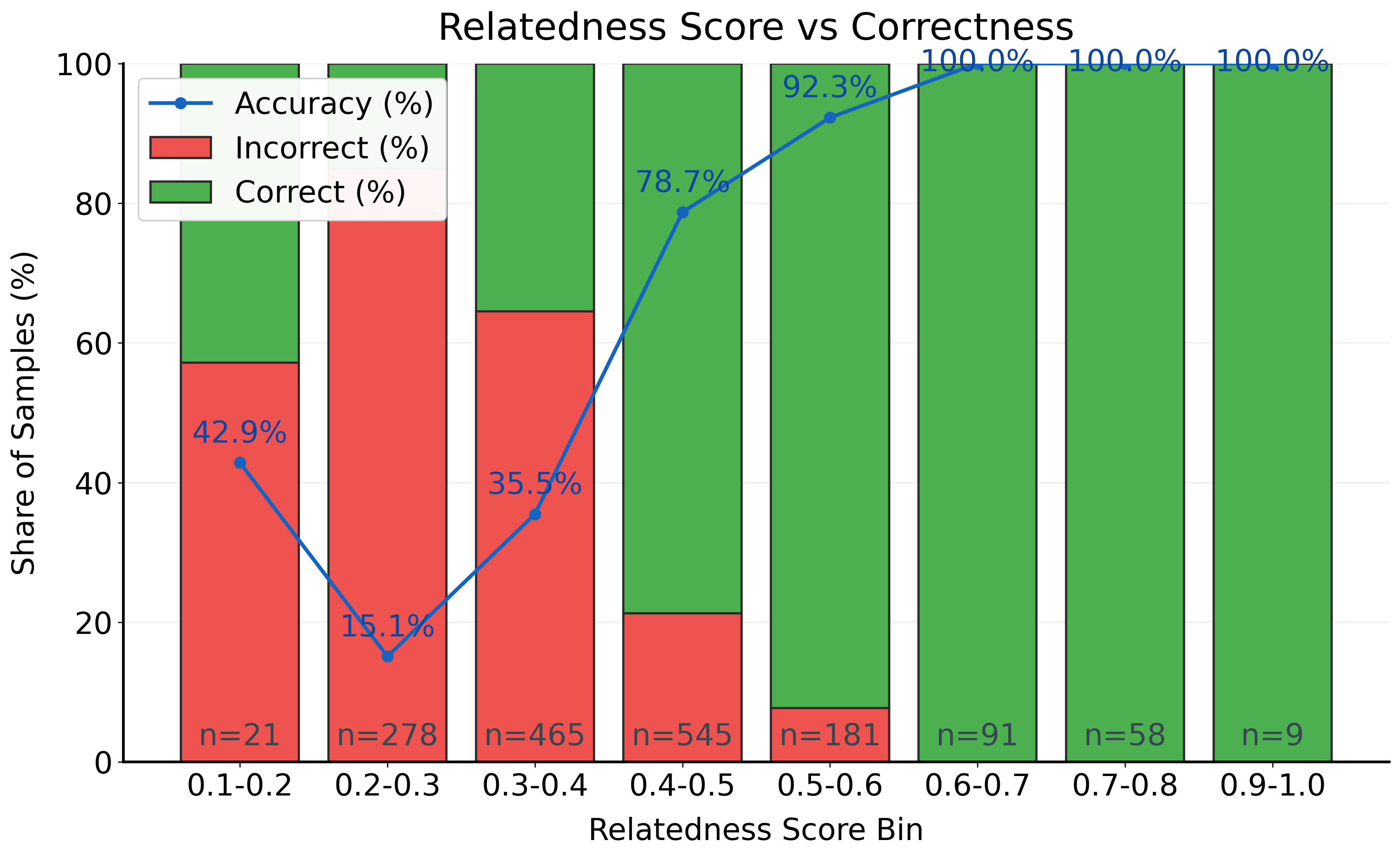}
    \caption{Percentages of correct and incorrect risk predictions across 8 relatedness score bins. 
    \textbf{Relatedness score} here refers to the average over all predicted items by all models for each video in our dataset; see Eq.~(\ref{rga_eq:1}). Correct predictions (green bars) exhibit significantly higher mean relatedness scores than incorrect ones (red bars), indicating that accurate risk anticipation is strongly tied to better visual grounding. 
    }
    \label{fig:relatedness_scores}
\end{figure}

There is a strong association between reasoning grounding and risk prediction accuracy. In Figure~\ref{fig:relatedness_scores}, we compute the average relatedness score (Eq.~\ref{rga_eq:1}) averaged over all predicted items by all the models for each video sample in our benchmark, and split the video samples into 8 relatedness score bins. As the relatedness score (measuring the alignment between model reasoning and visual features) increases, the proportion of correct predictions (green bar) rises substantially. Notably, the dip in the 0.2–0.3 relatedness bin, where accuracy is lowest (15.3\%), reveals a distinct failure mode in which the models generate ``hallucination'' reasoning that is not aligned with the true risk indicators in the image. In these cases, the models are not merely overlooking the risk; rather, they construct justification for their wrong answers based on irrelevant or misleading visual features.


\takeawaybox{
\textbf{VLM reasoning is poorly grounded in relevant risk objects.}  
Prediction accuracy strongly correlates with whether the model explicitly references relevant visual risk indicator objects.
}

\begin{figure}[!b]
    \centering
    \includegraphics[width=\linewidth]{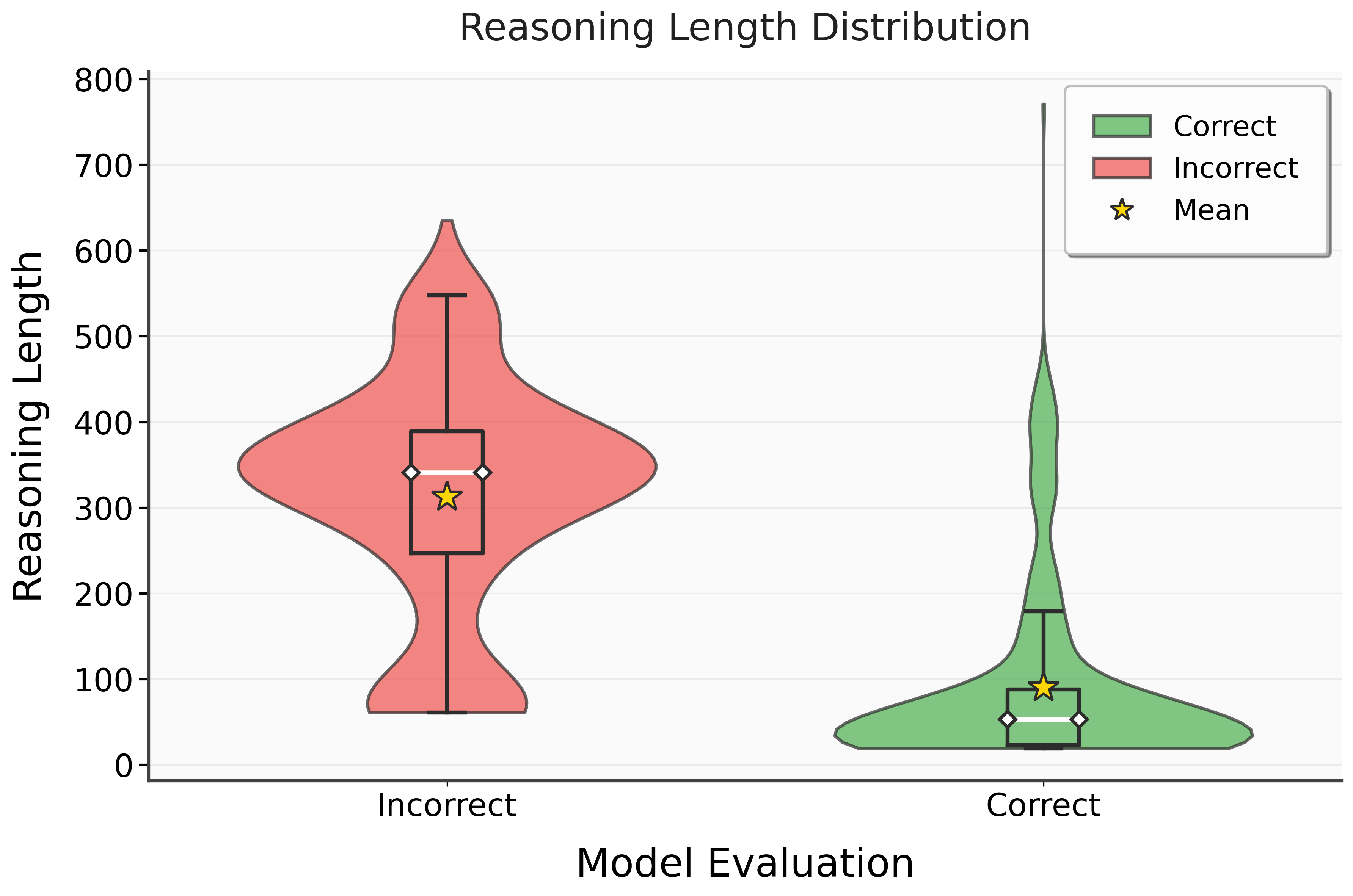}
    \caption{Reasoning length distribution across incorrect and correct predictions. Incorrect predictions (red) are associated with significantly longer and more complex reasoning traces compared to correct ones (green), suggesting that models' overthinking or circular deliberation often leads to performance degradation.}
    \label{fig:overthink}
\end{figure}
\subsubsection{Self-Correcting Reasoning Analysis.}
\label{sec:overthinking}

Models frequently engage in self-correction during risk prediction, as reflected by elevated confusion counts. However, self-correction consistently reduces accuracy by 15--26 percentage points across all experimental conditions as in Figure. \ref{fig:overthink}. Qualitative inspection reveals two dominant patterns: (1) rethinking fails to correct an initially wrong perception, and (2) rethinking overrides an initially correct judgment with speculative reasoning. This contrasts with prior work where deliberation improves performance, suggesting that uncertainty in early risk prediction stems from insufficient visual evidence rather than inadequate reasoning.

\takeawaybox{
\textbf{Overthinking degrades performance in risk prediction.}  
Unlike traditional reasoning tasks, additional deliberation consistently lowers accuracy.
}

\begin{figure*}[!ht]
    \centering
    \includegraphics[width=\linewidth]{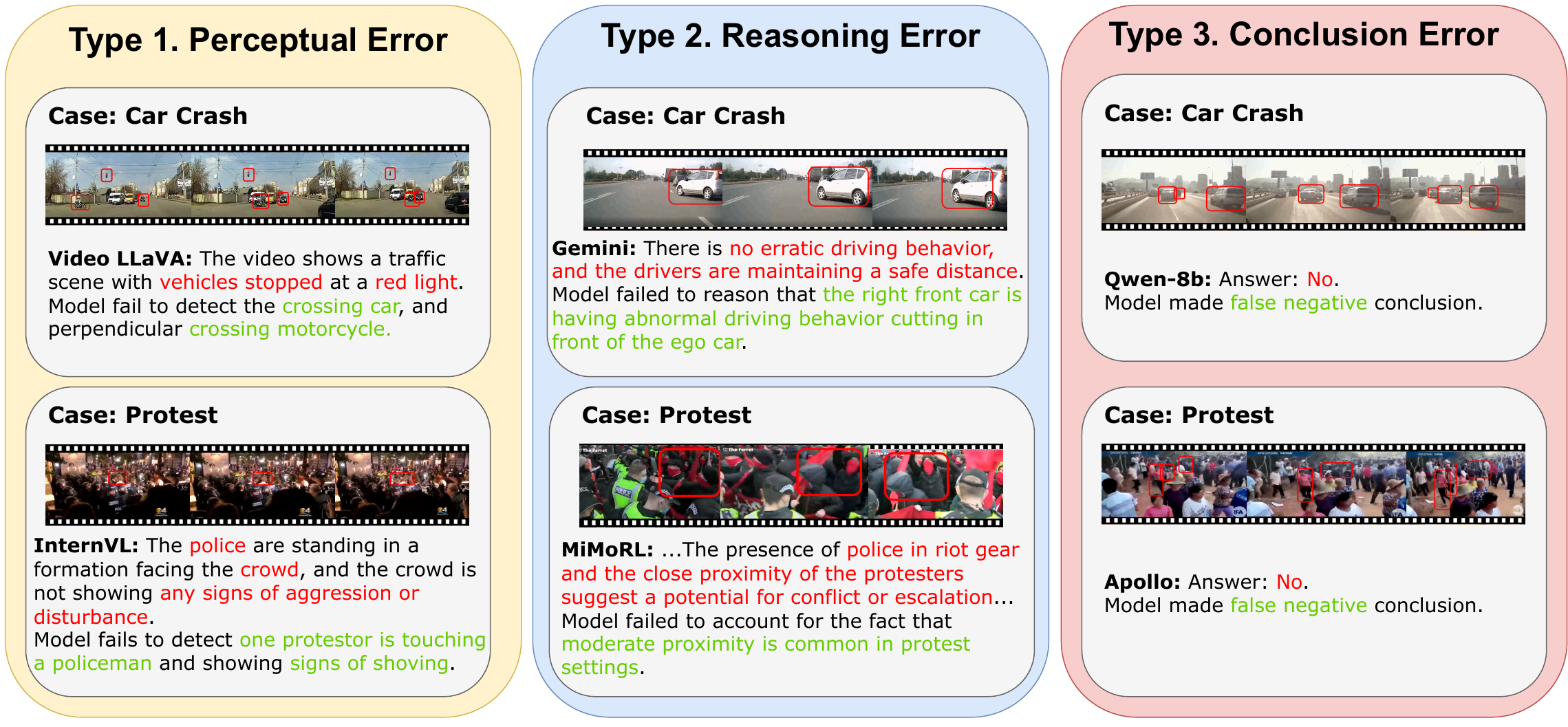}
    \caption{Taxonomy of VLM failure modes in situational risk assessment, categorized by perceptual, reasoning, and conclusion errors. Examples include perceptual errors (Type 1) that miss critical cues like aggressive shoving, reasoning errors (Type 2) like failing to interpret normal vs. abnormal behaviors, and incorrect answers (Type 3).}
    \label{fig:casestudy}
\end{figure*}

\subsubsection{Temporal Reasoning Analysis}
\label{sec:temporal_reasoning}
Table \ref{tab:temp_reasoning} illustrates the performance for each augmentation. Across domains, reversing or shuffling video frames results in only minor performance degradation (1--3\%). This suggests that VLM predictions are driven by the presence of salient frames rather than by temporal progression or causal ordering, revealing a lack of robust temporal reasoning. To ensure that this finding is not an artifact of sparse temporal sampling, we conduct an additional full-frame ablation on the Car Crash subset using 10 fps. As shown in Appendix \ref{tab:frame_sensitivity}, increasing the temporal resolution yields only marginal improvements in F1 and RGA, while TRD and SCD remain nearly unchanged, further supporting that current VLMs are limited by deficient temporal reasoning rather than by limited frame availability.


\takeawaybox{
    \textbf{Current VLMs lack genuine temporal reasoning.}  
    Performance differences between original, shuffled, and reversed video inputs are small, indicating that models rely primarily on static frame content rather than temporal order.
}

\begin{table}[h!]
\centering
\footnotesize
\resizebox{\columnwidth}{!}{
\begin{tabular}{lcccc}
\toprule
Video Type & Car Crash & Protest & Arson & Cruise Collision \\
\midrule
Basic       & $0.67 \pm 0.2$ & $0.59 \pm 0.2$ & $0.57 \pm 0.1$ & $0.62 \pm 0.2$ \\
Shuffled    & $0.68 \pm 0.1$ & $0.57 \pm 0.3$ & $0.59 \pm 0.1$ & $0.61 \pm 0.1$ \\
Swap Halves & $0.65 \pm 0.3$ & $0.58 \pm 0.3$ & $0.56 \pm 0.2$ & $0.59 \pm 0.3$ \\
Reversed    & $0.68 \pm 0.3$ & $0.58 \pm 0.3$ & $0.58 \pm 0.3$ & $0.62 \pm 0.3$ \\
\bottomrule
\end{tabular}
}
\caption{Model performance across temporal perturbations. Model performance is nearly identical across original and modified temporal sequences, suggesting that VLMs rely more on static frame content than genuine temporal reasoning for risk assessment.}
\label{tab:temp_reasoning}
\end{table}

\subsubsection{Effect of Risk Signal Length}
\label{sec:forecasting}
\begin{figure}[!ht]
    \centering
    \includegraphics[width=\linewidth]{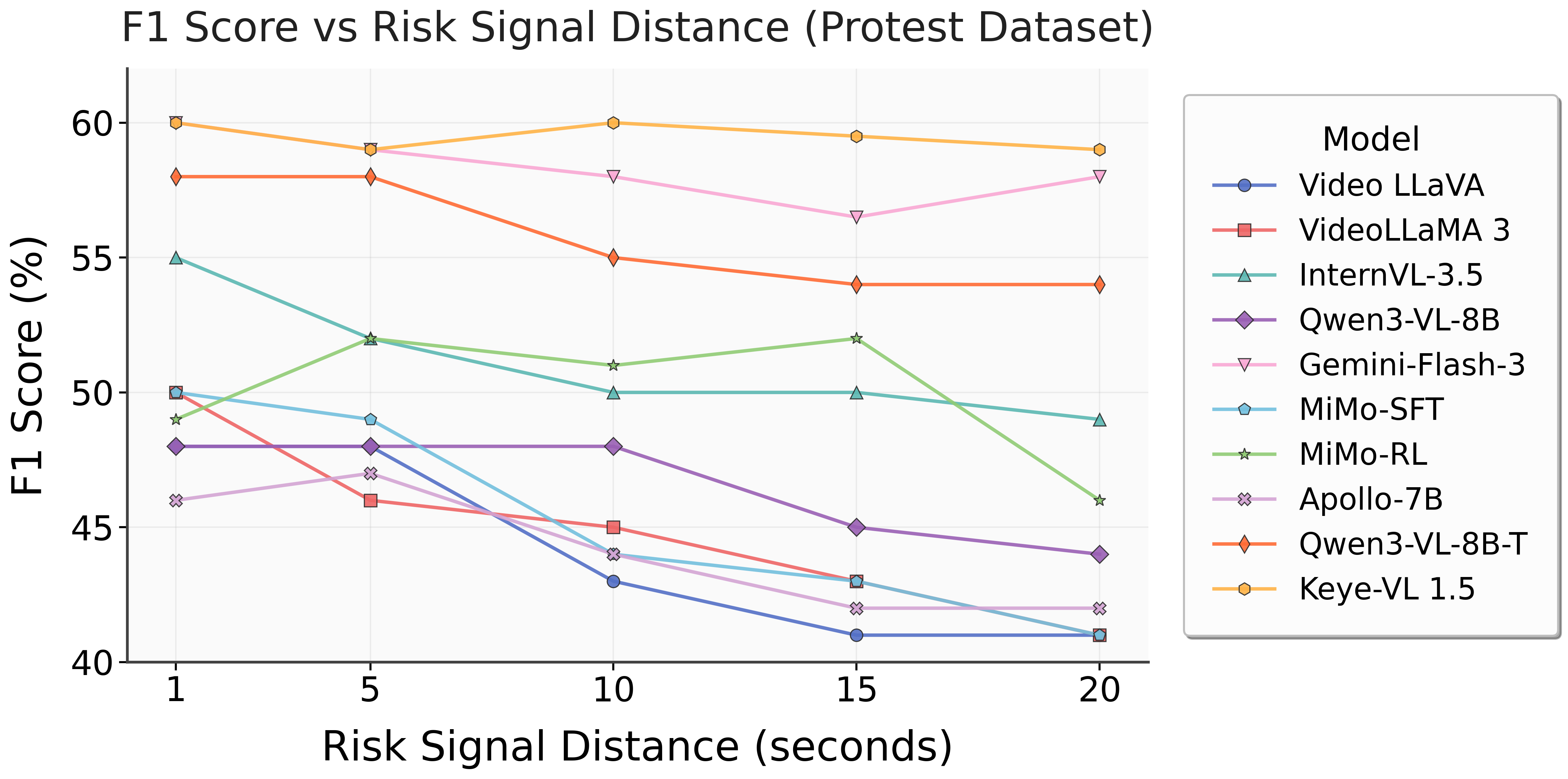}
    \caption{Model accuracy vs. Risk signal distance for Protest data. F1 scores decline as the temporal gap between the risk cue and the incident increases from 1 to 20 seconds, suggesting that VLMs struggle to maintain predictive accuracy when cues are temporally far.} 
    \label{fig:temporal}
\end{figure}

Aggregating results across the entire test set, we observed that: as the temporal gap between the risk signal and incident increases from 1 to 20 seconds,  
model accuracy systematically degrades, typically dropping from 45--50\% to 36--44\% . This shows that VLMs struggle to anticipate future risky events when predictive cues are subtle and temporally remote. Among evaluated models, MiMo-RL maintains stable performance across a temporal gap (1 to 20 sec lead time),
indicating improved robustness to early signals as in Figure~\ref {fig:temporal}.

\takeawaybox{
\textbf{Longer temporal distance weakens risk prediction.}  
Model accuracy declines as the risk signal becomes temporally distant from the incident.
}

\subsection{Qualitative Case Study: Error Analysis of Gemini on Car-Crash Videos}

To analyze VLM failures, we conducted a qualitative study of 336 incorrect predictions of best-performing model Gemini across domains. By manually inspecting reasoning traces, we derived an error taxonomy from recurring patterns (Figure~\ref{fig:casestudy}):

\begin{itemize}[leftmargin=13pt, itemsep=0pt, parsep=0pt] \item \textit{Perceptual Errors} (57.1\%): Missing or misinterpreting critical visual cues, such as aggressive gestures or unexpected vehicle entries. 
\item \textit{Reasoning Errors} (19.3\%): Failing to integrate detected cues, leading to causal misattribution or incomplete inference. 
\item \textit{Conclusion Errors}: Premature decisions, overconfidence, or internal contradictions. \end{itemize}

Perceptual errors were identified in 57.1\% of failures (192/336), while reasoning errors occurred in 19.3\% (65/336). These categories overlap: 9.2\% of failures contained both perceptual and reasoning errors. Perceptual failures were therefore the most common identifiable upstream source of incorrect predictions, often involving missed motion, trajectory, proximity, or behavioral cues. Reasoning errors instead reflected cases in which the model's own visual observations were sufficient to support an appropriate judgment, but those observations were converted into an incorrect risk inference. The complete overlap analysis is reported in Appendix \ref{app:error_overlap}.

\section{Conclusion}

We introduce a video benchmark \benchmark designed to evaluate models’ ability to predict real-life risk events, supported by a high-quality dataset constructed through model disagreement and LLM-based filtering. Beyond the benchmark itself, we develop a framework for identifying challenging scenarios and define metrics that capture temporal sensitivity, visual grounding, and reasoning behavior. Our evaluation shows that existing VLMs substantially lag behind human performance, relying on static visual cues for temporal reasoning and performing similarly under temporal perturbations. Moreover, models struggle to justify predicted risks through accurate visual grounding, and unlike in mathematics or coding, elaborate reasoning traces frequently degrade performance. We hope our benchmark, data curation pipeline, and evaluation code provide a foundation for evaluating VLMs in situational risk assessment.


\section*{Limitations}

While our study encompasses four diverse risk domains (protests, traffic incidents, arson, and cruise collisions), it does not cover all realistic hazards. \benchmark targets physically observable risks, omitting digital threats like misinformation or AI-generated content, leaving these for future study. Additionally, our evaluation pipeline relies on an LLM-based judge. While scalable and interpretable, it may occasionally hallucinate, introducing noise into the reported metrics. Finally, our temporal annotation framework requires precise manual labeling to pinpoint risk signal boundaries. Although crowdsourcing platforms like MTurk enabled scaling to our current benchmark size, the intense quality-control demands of frame-level annotation limit aggresive expansion. Future work will explore weakly-supervised or automated approaches to scale early risk localization further.

\section*{Ethical Considerations}
This research adheres to the ACL Code of Ethics. The data collection and annotation process was conducted under the following ethical considerations:

\paragraph{Data Collection and Privacy} 
The video data used in this benchmark were sourced from public platforms (YouTube) following the platform's Terms of Service. We have ensured that no private or personally identifiable information is explicitly highlighted or utilized beyond the scope of situational risk assessment.

\paragraph{Annotation Process} 
The annotation involves two phases: (1) a crowdsourced phase via MTurk (all participants paid above federal minimum wage), and (2) a benchmark phase performed by the authors. Research was conducted under an IRB-approved protocol to ensure ethical standards and avoid vulnerable populations. Using internal experts for reasoning traces ensured high quality control and domain alignment across protests, traffic incidents, arson, and cruise collisions. This hybrid approach balances the scalability needed for benchmark expansion with the domain expertise required for situational risk reasoning.


\paragraph{Potential Misuse} 
While \benchmark aims to improve public safety through early risk anticipation, we acknowledge that risk prediction models could potentially be used for unauthorized surveillance. We advocate for the use of this technology strictly within the bounds of legal and ethical safety-critical frameworks.

\bibliography{custom}

\begin{thebibliography}{23}
\providecommand{\natexlab}[1]{#1}

\bibitem[{Bai et~al.(2025)Bai, Chen, Liu, Wang, Ge, Song, Dang, Wang, Wang, Tang et~al.}]{bai2025qwen2}
Shuai Bai, Keqin Chen, Xuejing Liu, Jialin Wang, Wenbin Ge, Sibo Song, Kai Dang, Peng Wang, Shijie Wang, Jun Tang, and 1 others. 2025.
\newblock Qwen2. 5-vl technical report.
\newblock \emph{arXiv preprint arXiv:2502.13923}.

\bibitem[{Bao et~al.(2020)Bao, Yu, and Kong}]{bao2020uncertainty}
Wentao Bao, Qi~Yu, and Yu~Kong. 2020.
\newblock Uncertainty-based traffic accident anticipation with spatio-temporal relational learning.
\newblock In \emph{Proceedings of the 28th ACM International Conference on Multimedia}, pages 2682--2690.

\bibitem[{Caba~Heilbron et~al.(2015)Caba~Heilbron, Escorcia, Ghanem, and Carlos~Niebles}]{caba2015activitynet}
Fabian Caba~Heilbron, Victor Escorcia, Bernard Ghanem, and Juan Carlos~Niebles. 2015.
\newblock Activitynet: A large-scale video benchmark for human activity understanding.
\newblock In \emph{Proceedings of the ieee conference on computer vision and pattern recognition}, pages 961--970.

\bibitem[{Chen et~al.(2024)Chen, Liao, Lin, Yu, Chen, and Wang}]{NEURIPS2024_32683193}
Jr-Jen Chen, Yu-Chien Liao, Hsi-Che Lin, Yu-Chu Yu, Yen-Chun Chen, and Yu-Chiang~Frank Wang. 2024.
\newblock \href {https://doi.org/10.52202/079017-0900} {Rextime: A benchmark suite for reasoning-across-time in videos}.
\newblock In \emph{Advances in Neural Information Processing Systems}, volume~37, pages 28662--28673. Curran Associates, Inc.

\bibitem[{Chen et~al.(2025)Chen, Huang, Shi, Hu, Ye, Zhu, Liu, Molchanov, Kautz, Qi et~al.}]{chen2025scaling}
Yukang Chen, Wei Huang, Baifeng Shi, Qinghao Hu, Hanrong Ye, Ligeng Zhu, Zhijian Liu, Pavlo Molchanov, Jan Kautz, Xiaojuan Qi, and 1 others. 2025.
\newblock Scaling rl to long videos.
\newblock \emph{arXiv preprint arXiv:2507.07966}.

\bibitem[{Cheng et~al.(2025)Cheng, Hou, Tao, and Liao}]{cheng2025video}
Junhao Cheng, Liang Hou, Xin Tao, and Jing Liao. 2025.
\newblock Video-as-answer: Predict and generate next video event with joint-grpo.
\newblock \emph{arXiv preprint arXiv:2511.16669}.

\bibitem[{Fatima et~al.(2021)Fatima, Khan, and Kyung}]{fatima2021global}
Mishal Fatima, Muhammad Umar~Karim Khan, and Chong-Min Kyung. 2021.
\newblock Global feature aggregation for accident anticipation.
\newblock In \emph{2020 25th International conference on pattern recognition (ICPR)}, pages 2809--2816. IEEE.

\bibitem[{Fruhwirth-Reisinger et~al.(2025)Fruhwirth-Reisinger, Mali{\'c}, Lin, Schinagl, Schulter, and Possegger}]{fruhwirth2025stsbench}
Christian Fruhwirth-Reisinger, Du{\v{s}}an Mali{\'c}, Wei Lin, David Schinagl, Samuel Schulter, and Horst Possegger. 2025.
\newblock Stsbench: A spatio-temporal scenario benchmark for multi-modal large language models in autonomous driving.
\newblock \emph{arXiv preprint arXiv:2506.06218}.

\bibitem[{Hong et~al.(2025{\natexlab{a}})Hong, Cheng, Yang, Wang, Wang, Gu, Huang, Dong, and Tang}]{Hong_2025_CVPR}
Wenyi Hong, Yean Cheng, Zhuoyi Yang, Weihan Wang, Lefan Wang, Xiaotao Gu, Shiyu Huang, Yuxiao Dong, and Jie Tang. 2025{\natexlab{a}}.
\newblock Motionbench: Benchmarking and improving fine-grained video motion understanding for vision language models.
\newblock In \emph{Proceedings of the IEEE/CVF Conference on Computer Vision and Pattern Recognition (CVPR)}, pages 8450--8460.

\bibitem[{Hong et~al.(2025{\natexlab{b}})Hong, Yu, Gu, Wang, Gan, Tang, Cheng, Qi, Ji, Pan et~al.}]{hong2025glm}
Wenyi Hong, Wenmeng Yu, Xiaotao Gu, Guo Wang, Guobing Gan, Haomiao Tang, Jiale Cheng, Ji~Qi, Junhui Ji, Lihang Pan, and 1 others. 2025{\natexlab{b}}.
\newblock Glm-4.1 v-thinking: Towards versatile multimodal reasoning with scalable reinforcement learning.
\newblock \emph{arXiv preprint arXiv:2507.01006}.

\bibitem[{Hussain et~al.(2024)Hussain, Ali, Li, and Haque}]{hussain2024bi}
Fizza Hussain, Yasir Ali, Yuefeng Li, and Md~Mazharul Haque. 2024.
\newblock A bi-level framework for real-time crash risk forecasting using artificial intelligence-based video analytics.
\newblock \emph{Scientific Reports}, 14(1):4121.

\bibitem[{Kung et~al.(2024)Kung, Yang, Pao, Lu, Chen, Lu, and Chen}]{kung2024riskbench}
Chi-Hsi Kung, Chieh-Chi Yang, Pang-Yuan Pao, Shu-Wei Lu, Pin-Lun Chen, Hsin-Cheng Lu, and Yi-Ting Chen. 2024.
\newblock Riskbench: A scenario-based benchmark for risk identification.
\newblock In \emph{2024 IEEE International Conference on Robotics and Automation (ICRA)}, pages 14800--14807. IEEE.

\bibitem[{Li et~al.(2024)Li, Wang, He, Li, Wang, Liu, Wang, Xu, Chen, Luo et~al.}]{li2024mvbench}
Kunchang Li, Yali Wang, Yinan He, Yizhuo Li, Yi~Wang, Yi~Liu, Zun Wang, Jilan Xu, Guo Chen, Ping Luo, and 1 others. 2024.
\newblock Mvbench: A comprehensive multi-modal video understanding benchmark.
\newblock In \emph{Proceedings of the IEEE/CVF Conference on Computer Vision and Pattern Recognition}, pages 22195--22206.

\bibitem[{Lin et~al.(2023)Lin, Ye, Zhu, Cui, Ning, Jin, and Yuan}]{lin2023video}
Bin Lin, Yang Ye, Bin Zhu, Jiaxi Cui, Munan Ning, Peng Jin, and Li~Yuan. 2023.
\newblock Video-llava: Learning united visual representation by alignment before projection.
\newblock \emph{arXiv preprint arXiv:2311.10122}.

\bibitem[{Sultani et~al.(2018)Sultani, Chen, and Shah}]{sultani2018real}
Waqas Sultani, Chen Chen, and Mubarak Shah. 2018.
\newblock Real-world anomaly detection in surveillance videos.
\newblock In \emph{Proceedings of the IEEE conference on computer vision and pattern recognition}, pages 4771--4780.

\bibitem[{Swetha et~al.(2025)Swetha, Kuehne, and Shah}]{swetha2025timelogic}
Sirnam Swetha, Hilde Kuehne, and Mubarak Shah. 2025.
\newblock Timelogic: A temporal logic benchmark for video qa.
\newblock \emph{arXiv preprint arXiv:2501.07214}.

\bibitem[{Team et~al.(2025)Team, Yue, Lin, Song, Wang, Ren, Gu, Li, Li, Zhao, Li, Bao, Tian, Zhang, Wang, Zhu, Cici, He, Ye, Shen, Zhang, Jiang, Zheng, Song, Luo, Yu, Wang, Tian, Tu, Yan, Huang, Wang, Xu, Song, Zhang, Yong, Zhang, Deng, Yang, Ma, Lv, Zhuang, Liu, Deng, Liu, Chen, Yu, Liu, Wang, Ma, Wang, Wang, Chen, Zhu, Zhou, Zhou, Fang, Shi, Dong, Xiao, Xu, Liu, Xu, Qu, Zhao, Lv, Wang, Zhang, Zhang, Zhang, Ma, Liu, Cai, and Xia}]{coreteam2025mimovltechnicalreport}
Core Team, Zihao Yue, Zhenru Lin, Yifan Song, Weikun Wang, Shuhuai Ren, Shuhao Gu, Shicheng Li, Peidian Li, Liang Zhao, Lei Li, Kainan Bao, Hao Tian, Hailin Zhang, Gang Wang, Dawei Zhu, Cici, Chenhong He, Bowen Ye, and 55 others. 2025.
\newblock \href {https://arxiv.org/abs/2506.03569} {Mimo-vl technical report}.
\newblock \emph{Preprint}, arXiv:2506.03569.

\bibitem[{Wang et~al.(2025)Wang, Liu, Liu, Du, Kawaguchi, Wang, and Pang}]{wang2025fostering}
Haonan Wang, Hongfu Liu, Xiangyan Liu, Chao Du, Kenji Kawaguchi, Ye~Wang, and Tianyu Pang. 2025.
\newblock Fostering video reasoning via next-event prediction.
\newblock \emph{arXiv preprint arXiv:2505.22457}.

\bibitem[{Wu et~al.(2024)Wu, Yu, Chen, Tenenbaum, and Gan}]{wu2024star}
Bo~Wu, Shoubin Yu, Zhenfang Chen, Joshua~B Tenenbaum, and Chuang Gan. 2024.
\newblock Star: A benchmark for situated reasoning in real-world videos.
\newblock \emph{arXiv preprint arXiv:2405.09711}.

\bibitem[{Xun et~al.(2025)Xun, Tao, Li, Shi, Lin, Zhu, Yan, Li, Zhang, Wang et~al.}]{xun2025rtv}
Shuhang Xun, Sicheng Tao, Jungang Li, Yibo Shi, Zhixin Lin, Zhanhui Zhu, Yibo Yan, Hanqian Li, Linghao Zhang, Shikang Wang, and 1 others. 2025.
\newblock Rtv-bench: Benchmarking mllm continuous perception, understanding and reasoning through real-time video.
\newblock \emph{arXiv preprint arXiv:2505.02064}.

\bibitem[{Zhang et~al.(2025)Zhang, Li, Cheng, Hu, Yuan, Chen, Leng, Jiang, Zhang, Li et~al.}]{zhang2025videollama}
Boqiang Zhang, Kehan Li, Zesen Cheng, Zhiqiang Hu, Yuqian Yuan, Guanzheng Chen, Sicong Leng, Yuming Jiang, Hang Zhang, Xin Li, and 1 others. 2025.
\newblock Videollama 3: Frontier multimodal foundation models for image and video understanding.
\newblock \emph{arXiv preprint arXiv:2501.13106}.

\bibitem[{Zhou et~al.(2025)Zhou, Vilesov, He, Wan, Zhang, Nagachandra, Chang, Chen, Wang, and Kadambi}]{Zhou_2025_ICCV}
Shijie Zhou, Alexander Vilesov, Xuehai He, Ziyu Wan, Shuwang Zhang, Aditya Nagachandra, Di~Chang, Dongdong Chen, Xin~Eric Wang, and Achuta Kadambi. 2025.
\newblock Vlm4d: Towards spatiotemporal awareness in vision language models.
\newblock In \emph{Proceedings of the IEEE/CVF International Conference on Computer Vision (ICCV)}, pages 8600--8612.

\bibitem[{Zohar et~al.(2025)Zohar, Wang, Dubois, Mehta, Xiao, Hansen-Estruch, Yu, Wang, Juefei-Xu, Zhang et~al.}]{zohar2025apollo}
Orr Zohar, Xiaohan Wang, Yann Dubois, Nikhil Mehta, Tong Xiao, Philippe Hansen-Estruch, Licheng Yu, Xiaofang Wang, Felix Juefei-Xu, Ning Zhang, and 1 others. 2025.
\newblock Apollo: An exploration of video understanding in large multimodal models.
\newblock In \emph{Proceedings of the Computer Vision and Pattern Recognition Conference}, pages 18891--18901.

\end{thebibliography}
\newpage

\appendix
\section{Appendix}
\label{sec:appendix}


\subsection{Example List for Query Construction (Truncated for Readability)}
\label{app:query_construction}
As detailed in Section 3.1, we utilize a dual-stream curation strategy based on the nature of the domain. Table \ref{tab:macro_events} details the year-location anchoring used (e.g., protest). Table \ref{tab:micro_events} details the combined vocabulary matrices used to extract raw footage for localized micro-anomalies (e.g., arson), explicitly filtering out abstracted news broadcasts.

\begin{table*}[t]
\small
\renewcommand{\arraystretch}{1.15}
\begin{tabularx}{\textwidth}{c l X c X}
\toprule
\textbf{Year} & \textbf{Location} & \textbf{Specific Cities} & \textbf{Domain Synonyms} & \textbf{Example Query} \\
\midrule

\multicolumn{5}{c}{\textbf{Domain: Protests}} \\
\midrule
2000 & Palestine & Ramallah, Nablus, Hebron, \ldots & 
\multirow{10}{*}{\parbox{3.2cm}{\centering civil unrest,\\ protest,\\ rally,\\ sit-in,\\ strike,\\ march,\\ demonstration}} & 
2000 Palestine civil unrest; 2000 Ramallah civil unrest\dots \\

2003 & Global & New York City, London, Tokyo, \ldots & & 
2003 Global civil unrest; 2003 New York City march\dots \\

2011 & Egypt & Cairo, Alexandria, Suez, \ldots & & 
2011 Cairo demonstration\dots \\

2014 & Hong Kong & Hong Kong & & 
2014 Hong Kong strike\dots \\

2020 & United States & Minneapolis, New York City, \ldots & & 
2020 Minneapolis civil unrest\dots \\

2024 & India & New Delhi, Amritsar, Chandigarh, \ldots & & 
2024 New Delhi protest\dots \\

\bottomrule
\end{tabularx}
\caption{Example List for Macro-Year-Location Query Construction}
\label{tab:macro_events}
\end{table*}

\begin{table*} 
\centering
\small
\renewcommand{\arraystretch}{1.2}

\begin{tabularx}{\textwidth}{l l X}
\toprule
\textbf{Domain} & \textbf{Vocabulary Group} & \textbf{Included Terms \& Constraints} \\
\midrule

\multirow{5}{*}{\textbf{Arson}} 
& \textbf{Evidence Terms} & caught on camera, caught on tape, CCTV , surveillance video, security camera, doorbell camera, dashcam, footage \\
& \textbf{Action Terms} & set on fire, setting fire, pours gasoline, pouring petrol, lights himself on fire, sets ablaze, torching \\
& \textbf{Actor Terms} & suspect, arsonist, man, masked, criminal, blundering, wanted \\
& \textbf{Target Terms} & car, truck, vehicle, house, restaurant, store, church, mosque, gas station, strip mall, cop car, apartments \\
\cmidrule{2-3}
& \textit{Example Query} & \textit{(surveillance video) AND (pours gasoline) AND (masked suspect) AND (``gas station'')} \\

\midrule

\multirow{4}{*}{\textbf{Cruise}} 
& \textbf{Base Terms} & cruise ship collision, passenger ship collision, ferry collision, maritime collision, vessel collision, cruise ship hits dock, ship docking accident, bridge strike ship \\
& \textbf{Video Terms} & raw footage, caught on camera, CCTV, security footage, witness video, bystander video, bridge cam, port cam \\
& \textbf{Context Terms} & dock, pier, port, harbor, terminal, canal \\
\cmidrule{2-3}
& \textit{Example Query} & \textit{(cruise ship hits dock) AND (witness video) AND (terminal)} \\

\bottomrule
\end{tabularx}
\caption{Example List for Micro-Anomaly Combination Queries (Arson and Cruise Collisions)}
\label{tab:micro_events}
\end{table*}
\FloatBarrier

\clearpage

\begin{table*}[t]
\centering
\small

\subsection{Visual Quality and Authenticity Scoring Criteria}
\label{tab:quality_scoring}
\renewcommand{\arraystretch}{1.3}
\begin{tabularx}{\textwidth}{c l X X X X}
\toprule
\textbf{No.} & \textbf{Dimension} & \textbf{Criterion} & \textbf{Score 0} & \textbf{Score 1} & \textbf{Score 2} \\
\midrule
1 & Logo Assessment & Video should have no news organization branding or features &
Obvious news logos, watermarks, or media branding visible &
Ambiguous logos or possible reposts with unclear branding &
No branding visible; appears bystander- or CCTV-style \\

2 & Location Information & Video should contain no location mentioned in text or audio &
Explicit location references (city, state, country) &
Ambiguous or hypothetical location references &
No location information present \\

3 & Time/Date Information & Video should contain no time or date mentioned in text or audio &
Explicit time/date references (e.g., dates, timestamps) &
Ambiguous or unrealistic temporal references &
No time/date information present \\

4 & Reporter Presence & Video should include no reporter, anchor, or journalist &
Clearly identifiable reporter or journalist present &
Possibly media-affiliated speaker or unclear role &
No formal speaker; only participants or bystanders \\

5 & SNS Engagement Overlays & Video should contain no social media overlays or engagement icons &
Likes, shares, emojis, or UI elements visible &
Minimal or transient overlays &
No overlays or engagement elements \\

6 & Natural Image Quality & Video should show no signs of professional editing &
Heavy editing, stylized transitions, montage effects &
Some editing artifacts or unclear transitions &
Continuous, shaky, or natural camera movement \\

7 & Natural Temporal Continuity & Video should be a continuous recording &
Obvious jump cuts, stitched scenes, or time gaps &
Possible continuity, but breaks unclear &
One uninterrupted shot with natural temporal flow \\

8 & Consequence Text & Video should include no embedded text describing consequences &
Text describing severity (arrests, crackdowns, deaths) &
Minor or unclear commentary text &
No consequence text present \\

9 & Title / Description / Banner Text & Video should have no inflammatory or descriptive banners &
Mentions violence, arrests, or protest names &
Vague or unclear text &
Only factual metadata (e.g., place, date) \\

10 & Subtitle Text & Video should contain no speech transcription subtitles &
Subtitles clearly present &
Partial or unclear subtitle presence &
No subtitles visible \\

11 & Camera Perspective & Video should appear from participant, bystander, or CCTV view &
Professional media vantage point &
Drone or distant zoomed footage &
Ground-level, handheld, immersed, or surveillance view \\

12 & Post-Production Effects & Video should have no post-processing effects &
Filters, slow motion, stabilization effects visible &
Minor or uncertain processing &
Completely raw, unfiltered footage \\
\bottomrule
\end{tabularx}
\caption{Visual Quality and Authenticity Scoring Criteria}
\label{tab:visualquality}
\end{table*}

\FloatBarrier

\clearpage
\begin{table*}[t]
\centering
\small

\subsection{Examples of Risk and Non-Risk Annotations Across Domains}
\label{app:annotation_examples}

\renewcommand{\arraystretch}{1.1}
\begin{tabularx}{\textwidth}{l X X X X}
\toprule
\textbf{Label} & \textbf{Protest (Risk)} & \textbf{Protest (No-Risk)} & \textbf{Arson (Risk)} & \textbf{Arson (No-Risk)} \\
\midrule
Risk Signal Start & 00:03:00 & 00:04:00 & 00:01:15 & 00:02:00 \\
Risk Signal End & 00:12:58 & 00:08:55 & 00:08:12 & 00:05:00 \\
Risk Visual Indicator & Throwing gestures, aggressive body postures, raised batons, riot shields.  & Singing, dancing, sign-holding, and stationary police lines.  & Plastic bottle, dousing liquid from the bottle, lighter. & Birthday cake, candles on cake, parent monitoring. \\
Risk Signal Description & First police vehicles appear and protestors intend to throw objects; then officers raise baton toward protesters. 
 & Protesters hold signs still, some sing and dance while police observe from a distance without engagement.  & First a man in black hoodie enter the room carrying a big plastic bottle, then he douse liquid all over the room from the bottle, afterward he takes out a lighter from pocket. & First a girl shows up sitting in front of a birthday cake, next her parents hold her hand to light up the candle on the cake with a burning match. \\
Accident Start Frame & 00:12:59 & / & 00:08:13 & / \\
Accident End Frame & 00:15:00 & / & 00:12:00 & / \\
Accident Description & A smoke device is deployed into the crowd, causing protesters to disperse. & / & House interior ignites rapidly. & / \\
Risk Label & Yes & No & Yes & No \\
\bottomrule
\end{tabularx}
\caption{Annotation Examples: Protests \& Arson}
\label{tab:annotation_behavioral}
\end{table*}

\begin{table*}[t]
\centering
\small
\renewcommand{\arraystretch}{1.1}
\begin{tabularx}{\textwidth}{l X X X X}
\toprule
\textbf{Label} & \textbf{Car Crash (Risk)} & \textbf{Car Crash (No-Risk)} & \textbf{Cruise (Risk)} & \textbf{Cruise (No-Risk)} \\
\midrule
Risk Signal Start & 00:05:12 & 00:39:38 & 00:02:45 & 00:10:00 \\
Risk Signal End & 00:10:11 & 00:44:00 & 00:07:30 & 00:15:00 \\
Risk Visual Indicator & Crossing vehicle, red light, arrow on the ground indicating traffic flow. 
 &  Lane mark, steady speed, stop sign & Large cruise ship moving fast toward the dock, loud warning horn sounding continuously, people on the dock starting to run away in panic, ship appears too large to stop in time. & Slow controlled approach to port, steady movement, maintaining course toward the docking area, no visual signs of impact or distress. \\
Risk Signal Description & First a car enters the intersection against traffic flow; then ego continues when traffic light is red. 
 & Vehicles slow appropriately at stop sign and proceed steadily according to traffic signals within lane mark correctly. & First, the cruise ship approaches the dock at a high speed with its horn blaring. Then, it becomes clear it cannot stop as people on the pier begin to scatter. Afterwards, the ship maintains its trajectory directly toward the stationary river boat and dock. & First, the cruise ship approached the pier slowly. Then, it maintained a steady path for docking. Afterwards, it successfully reached the designated area without any issues. \\
Accident Start Frame & 00:10:12 & / & 00:07:31 & / \\
Accident End Frame & 00:13:00 & / & 00:11:00 & / \\
Accident Description & Following a collision, the vehicle spins and stops near the roadside. 
 & / & The cruise ship collided with a smaller river boat and scraped along the dock, causing significant damage and panic among bystanders. & / \\
Risk Label & Yes & No & Yes & No \\
\bottomrule
\end{tabularx}
\caption{Annotation Examples: Car Crash \& Cruise Collisions}
\label{tab:annotation_kinematic}

\end{table*}
\FloatBarrier

\clearpage
\begin{figure*}[ht]
    \subsection{Prompt Used for Quality Filtering of 12 Selection Criteria.}
    \label{12critera_prompt}
    \begin{AIbox}{Selection Criteria}
    \scriptsize
    \centering
    \begin{minipage}{0.95\linewidth} 
    \vspace{5pt}
    You are a video analysis expert tasked with evaluating YouTube videos based on 12 specific criteria. Analyze the provided video and assign scores for each category based on the detailed scoring rubric below.

    \textbf{SCORING SYSTEM} \\
    For each category, assign exactly one score: \\
    Score 0 = Fails to meet criterion \\
    Score 1 = Partially meets or unclear \\
    Score 2 = Fully meets criterion

    \textbf{EVALUATION CRITERIA}

    \textbf{1. Logo Assessment} \\
    Criterion: Video should have no news organization branding or features \\
    Score 0: Obvious news marks, logos, or clear media branding visible \\
    Score 1: Ambiguous news marks/logos or possible reposts with unclear branding \\
    Score 2: No branding visible, appears to be bystander-style or CCTV-style capture

    \textbf{2. Location Information} \\
    Criterion: Video should have no location mentioned in text or audio \\
    Score 0: Obvious location references in text or audio (city, state, country) \\
    Score 1: Ambiguous location references (hypothetical/unrealistic places) \\
    Score 2: No location information present in text or audio

    \textbf{3. Time/Date Information} \\
    Criterion: Video should have no time/date mentioned in text or audio \\
    Score 0: Obvious time/date references (e.g., "May 25th, 2025", "13:00pm") \\
    Score 1: Ambiguous temporal references (e.g., "a thousand years ago/later") \\
    Score 2: No time/date information present in text or audio

    \textbf{4. Reporter Presence} \\
    Criterion: Video should have no reporter, anchor, or journalist present \\
    Score 0: Reporter, journalist, or anchor clearly present and identifiable \\
    Score 1: Possibly media-affiliated speaker or unclear professional presence \\
    Score 2: No formal speaker present, only participants/bystanders

    \textbf{5. SNS Engagement Overlays} \\
    Criterion: Video should have no social media overlays or engagement icons \\
    Score 0: Social media overlays/pop-ups clearly visible (likes, shares, emojis) \\
    Score 1: Minimal or transient overlays present \\
    Score 2: No overlays or engagement elements visible

    \textbf{6. Natural Image Quality} \\
    Criterion: Video should show no signs of professional editing \\
    Score 0: Highly edited with stylized transitions, cuts, or montage effects \\
    Score 1: Some editing artifacts present or unclear transitions \\
    Score 2: Continuous, shaky, or natural camera movement with no professional editing

    \textbf{7. Natural Temporal Continuity} \\
    Criterion: Video should be continuous recording without interruptions \\
    Score 0: Obvious jump cuts, stitched scenes, or time gaps \\
    Score 1: Possibly continuous but breaks/transitions unclear \\
    Score 2: One uninterrupted shot with natural temporal flow

    \textbf{8. Consequence Text} \\
    Criterion: Video should have no embedded text about protest consequences \\
    Score 0: Text describing consequences/severity (arrests, crackdowns, deaths) \\
    Score 1: Unclear or minor commentary text present \\
    Score 2: No consequence text; only basic metadata like location/time allowed

    \textbf{9. Title/Description/Banner Text} \\
    Criterion: Video should have no inflammatory title/description banners \\
    Score 0: Text mentioning violence, specific protest names, arrests \\
    Score 1: Title/description is unclear or vague \\
    Score 2: Only factual metadata like place/date present

    \textbf{10. Subtitle Text} \\
    Criterion: Video should have no speech transcription subtitles \\
    Score 0: Subtitles clearly present showing speech transcriptions \\
    Score 1: Some subtitle presence but hard to distinguish \\
    Score 2: No subtitles visible at all

    \textbf{11. Camera Perspective} \\
    Criterion: Video should appear to be from participant, bystander, or CCTV perspective \\
    Score 0: Clearly taken from media zone or professional media position \\
    Score 1: Drone footage or distant zoomed footage \\
    Score 2: Ground-level, hand-held, immersed in crowd, or surveillance camera

    \textbf{12. Post-Production Effects} \\
    Criterion: Video should have no post-processing effects applied \\
    Score 0: Clear post-processing (filters, slow motion, stabilization effects) \\
    Score 1: Some smoothing effects or unknown processing present \\
    Score 2: Completely raw, unfiltered footage

    \textbf{OUTPUT FORMAT} \\
    Provide your analysis as a JSON array containing exactly 12 objects, one for each category. Each object must include: \\
    category\_number: integer (1-12) \\
    category\_name: string (exact name from criteria above) \\
    score: integer (0, 1, or 2)

    \textbf{ANALYSIS INSTRUCTIONS} \\
    1. Watch the entire video carefully \\
    2. Listen to all audio content \\
    3. Examine all visible text and overlays \\
    4. Assess video quality and editing characteristics \\
    5. Score each category independently \\
    6. If uncertain between two scores, choose the lower score \\
    7. Ensure all 12 categories are evaluated and included in response
    \vspace{5pt}
    \end{minipage}
    \end{AIbox}
    \label{tab:evaluation_prompts}
\end{figure*}

\FloatBarrier

\clearpage
\begin{figure*}[t]
    \subsection{Prompt Used for Reasoning Chain Evaluation}
    \begin{AIbox}{Reasoning Trace Evaluation Prompt}
    \scriptsize
    \centering
    \begin{minipage}{0.96\linewidth}
    \vspace{5pt}
    You are analyzing a reasoning chain from Video-Language Models to evaluate its quality and characteristics. Please analyze the following reasoning chain and provide the requested metrics.
    
     {GROUND TRUTH:} \{ground\_truth\} \\
     {MODEL PREDICTION:} \{model\_pred\} \\
    {REASONING CHAIN:} \{reasoning\}\\

    \textbf{Please analyze this reasoning chain and provide:}

    \textbf{1. Confusion Count:} Count the number of times the reasoning shows confusion, uncertainty, or self-correction. Look for phrases like:
    \begin{itemize}[noitemsep,topsep=2pt,leftmargin=15pt]
       \item "wait..."
       \item "no let me think again"
       \item "actually..."
       \item "hold on..."
       \item "let me reconsider"
       \item "I'm confused"
       \item "that doesn't seem right"
       \item Similar expressions of uncertainty or backtracking
    \end{itemize}

    \textbf{2. Decision Items:} Extract ALL items, objects, or things that the model specifically mentions from the video as part of its reasoning process for reaching its conclusion. This includes:
    \begin{itemize}[noitemsep,topsep=2pt,leftmargin=15pt]
       \item Physical objects mentioned in the reasoning
       \item Items that influenced the decision (whether risky or safe)
       \item Specific things the model identified or considered
       \item Objects that were part of the analysis
       \item Any concrete items/things mentioned that contributed to the final decision
       \item Both safe items (for non-risk scenarios) and dangerous items (for risk scenarios)
    \end{itemize}

    \textbf{OUTPUT FORMAT} \\
    Provide your analysis in the requested JSON format.
    \vspace{5pt}
    \end{minipage}
    \end{AIbox}
\end{figure*}
\FloatBarrier

\clearpage
\subsection{Risk Signal Temporal Distribution}

\noindent


\begin{figure*}[h!]
    \centering
    \includegraphics[width=\linewidth]{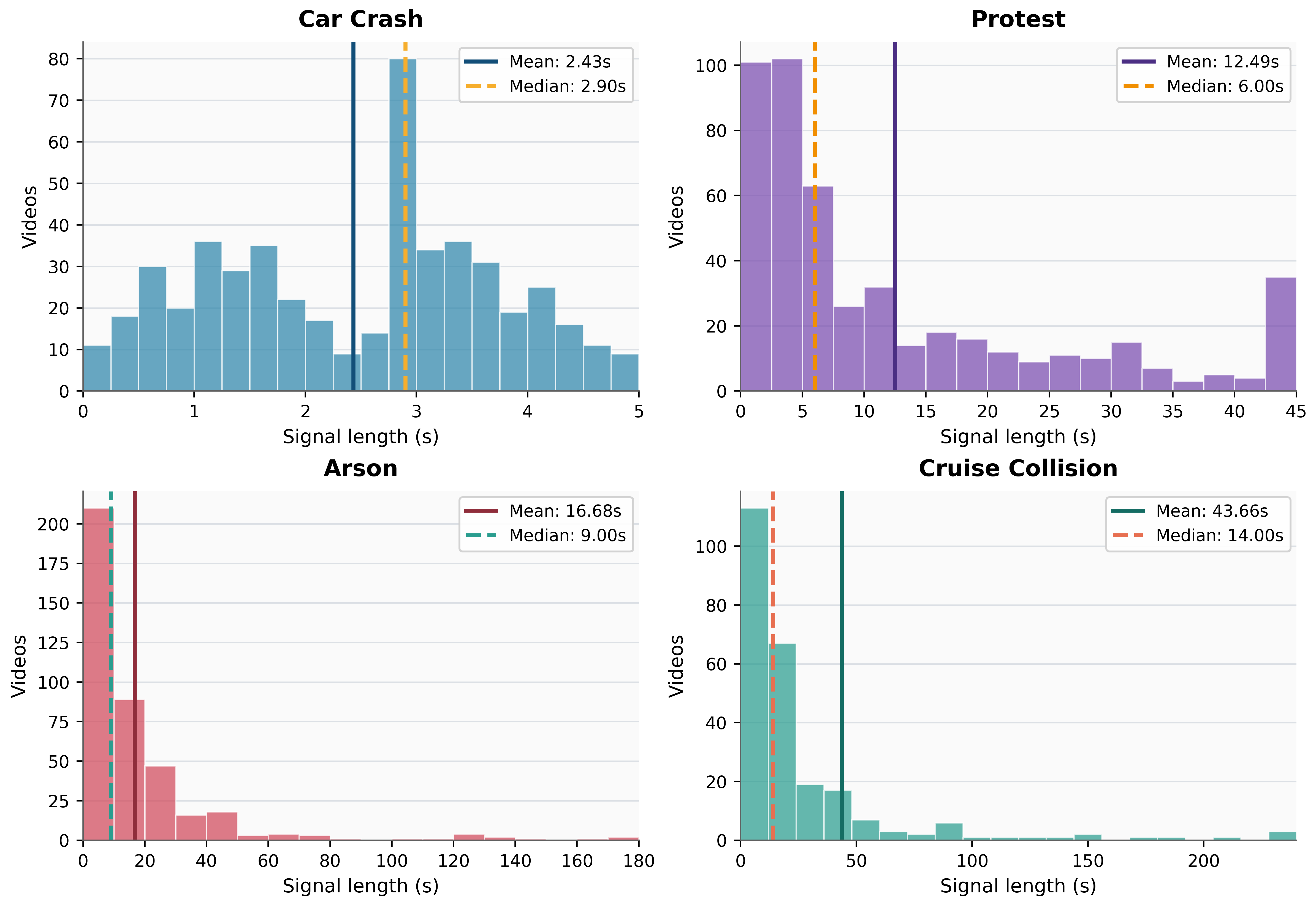}
    \caption{The histograms illustrate the distribution of risk signal lengths across four visual risk domains: car crashes, protests, arson, and cruise collisions. A ``risk signal'' is defined as the observable temporal window before an incident or outcome during which visible cues of potential danger are present.}
    \label{fig:combined_risk_length}
\end{figure*}
 
    

    In Figure~\ref{fig:combined_risk_length}, we report the histograms that illustrate the distribution of risk signal lengths across four visual risk domains: car crashes, protests, arson, and cruise collisions. 
    
    \textbf{Protest Dataset.} The data shows a significantly broader distribution of signal lengths. The mean duration is 12.5s, while the median is 6.0s. This suggests that risk in protest scenarios often involves a gradual escalation, with some signals extending beyond 40 seconds.
    
    \vspace{5pt}
    
    \textbf{Car Crash Dataset.} In contrast, the risk signals for vehicular accidents are highly concentrated and brief. The mean length is 2.4s and the median is 2.9s, indicating that the critical decision-making window for safety systems in these scenarios is very narrow.

    \vspace{5pt}

    \textbf{Arson Dataset.} The arson dataset is strongly right-skewed, with many short signals and a smaller number of much longer cases. The mean duration is 16.68s, while the median is 9.00s. This indicates that arson-related risk is often signaled by relatively short visual cues, although some incidents involve extended pre-event activity.
    
    \vspace{5pt}
    
    \textbf{Cruise Collision Dataset.} Risk signals for cruise collisions are substantially more protracted and variable. The mean signal length is 43.66s and the median is 14.00s, suggesting that these incidents often involve longer developmental phases than the other domains. However, the gap between the mean and median also indicates a long-tailed distribution driven by a subset of very long signal windows.

\FloatBarrier

\clearpage


\subsection{Representative Failure Cases}
\label{sec:qualitative_cases}
We present representative failure cases from Gemini to illustrate how perceptual and reasoning errors manifest in practice in Table~\ref{tab:errors}. These examples are selected from the car-crash domain and are intended to highlight common patterns rather than exhaustively enumerate all error types.

\begin{table*}[h!]
\hrule 
\vspace{1.5em}

\begin{multicols}{2}
\paragraph{Perceptual Errors}\mbox{}\\
\textbf{Case 1: Snow-caused loss of control misperceived as a normal intersection.} \\ 
\textit{Annotation.} ``First, a black car on the opposing lane slips on snow, then it drifts out of control.''  \\
\textit{Model output.} Gemini describes a routine urban intersection with functioning traffic lights and a yellow taxi.   
\textit{Diagnosis.} The model fails to indicate the hazardous road condition of snow and the resulting loss of traction, instead substituting an unrelated scene with different objects and traffic structure. The core visual indicators are entirely absent from the model’s perception.

\textbf{Case 2: Wrong-way driving replaced by an empty parking garage.}\\
\textit{Annotation.} ``First, the car in front drives in the opposing direction, then it continues into the driver’s lane.''  \\
\textit{Model output.} The model describes an empty underground parking garage with no moving vehicles. \\ 
\textit{Diagnosis.} The active roadway, opposing traffic flow, and dynamic agents are replaced by a static and unrelated environment.

\paragraph{Reasoning Errors}\mbox{}\\
\textbf{Case 3: ``No collision yet'' interpreted as no risk.  }\\
\textit{Annotation.} ``The white car in front brakes suddenly while the ego vehicle is still approaching.'' \\ 
\textit{Model output.} Gemini concludes that no risk is present because traffic lights are functioning and no crash is visible.  \\
\textit{Diagnosis.} Although the sudden braking event is acknowledged, the model incorrectly equates the absence of an observed collision with safety, ignoring the temporal consequence of risk.

\textbf{Case 4: Normal lane change framed as risky via hypothetical hazards. \\}
\textit{Annotation.} ``Ego vehicle changes to the right lane at normal speed.'' \\ 
\textit{Model output.} The model predicts risk by invoking slippery roads or poor weather conditions not observed.  \\
\textit{Diagnosis.} The model introduces speculative weather hazards instead of reasoning based on the described driving behavior, applying a generic safety heuristic without causal grounding of visual indicators in the video.

\paragraph{Conclusion Errors}\mbox{}\\
\textbf{Case 5: Collision labeled as no risk.} \\ 
\textit{Annotation.} ``First, a black car in front signals left and turns, at the same time it does not yield to the ego vehicle.''  \\
\textit{Model output.} Gemini predicts no risk. \\ 
\textit{Diagnosis.} The cars collide after the signal, while the model assigns a negative risk label.

\textbf{Case 6: Benign stop at red light labeled as risky.}  \\
\textit{Annotation.} ``During the red light, vehicles are waiting, and vehicles going straight from the right intersection are moving at normal speed the whole time.''  \\
\textit{Model output.} Gemini predicts a risk. \\ 
\textit{Diagnosis.} The model incorrectly flags a compliant and stationary driving scenario as risky.
\end{multicols}
\vspace{1em}
\hrule
\caption{Case studies for different error types}
\label{tab:errors}
\end{table*}
\FloatBarrier

\clearpage

\begin{table*}[h!]
    \centering

    \subsection{Temporal Resolution Ablation with Full-Frame Sampling}
    \label{tab:frame_sensitivity}

    \small
    \resizebox{\textwidth}{!}{
    \begin{tabular}{llcccc}
    \toprule
    \textbf{Model Category} & \textbf{Model} & \textbf{F1} & \textbf{RGA} & \textbf{TRD} & \textbf{SCD} \\
    \midrule
    \multirow{5}{*}{Non-Reasoning} 
        & Video-LLaVA    & $0.54 \pm 0.05$ & $49.6 \pm 5.48$ & $1.86 \pm 0.63$ & - \\
        & VideoLLaMA 3   & $0.60 \pm 0.04$ & $55.4 \pm 6.18$ & $1.63 \pm 0.58$ & - \\
        & Qwen3-VL-8B    & $0.65 \pm 0.04$ & $58.7 \pm 5.09$ & $1.51 \pm 0.52$ & - \\
        & MiMo-SFT       & $0.59 \pm 0.05$ & $52.8 \pm 7.31$ & $2.09 \pm 0.74$ & - \\
        & Apollo-7B      & $0.58 \pm 0.06$ & $54.9 \pm 6.77$ & $1.85 \pm 0.61$ & - \\
    \midrule
    \multirow{5}{*}{Reasoning}
        & InternVL-3.5     & $0.62 \pm 0.05$ & $65.7 \pm 4.91$ & $2.89 \pm 0.68$ & $-0.27 \pm 0.05$ \\
        & Qwen3-VL-8B-T    & $0.67 \pm 0.04$ & $70.1 \pm 4.19$ & $2.24 \pm 0.57$ & $-0.31 \pm 0.04$ \\
        & MiMo-RL          & $0.64 \pm 0.06$ & $62.8 \pm 6.88$ & $3.71 \pm 0.82$ & $-0.33 \pm 0.07$ \\
        & Keye-VL 1.5      & $0.58 \pm 0.05$ & $65.2 \pm 5.51$ & $3.17 \pm 0.71$ & $-0.22 \pm 0.04$ \\
        & GLM 4.1 V-T      & $0.59 \pm 0.04$ & $66.5 \pm 5.24$ & $3.36 \pm 0.76$ & $-0.28 \pm 0.05$ \\
    \midrule
    \multirow{2}{*}{Baselines}
        & Random Guess   & $0.50$ & - & - & - \\
        & Human          & $0.98 \pm 0.01$ & - & - & - \\
    \bottomrule
    \end{tabular}}
    \caption{Follow-up experiment on the Car Crash scenario using all 50 frames. Increasing the frame rate yields only marginal improvements in F1 and RGA, while TRD and SCD remain largely unchanged.}
    \label{tab:car_crash_50_frames}
\end{table*}

\FloatBarrier

\clearpage

\subsection{Human Annotation and Evaluation Protocol}
\label{app:annotation}

\subsubsection{Use of Annotation}

RiskCueBench uses human annotations for three distinct purposes:

\begin{enumerate}
    \item \textbf{Binary risk classification:} establishing the ground-truth risk/no-risk labels used for model evaluation.
    \item \textbf{Temporal localization:} identifying the risk signal clip, defined as the earliest temporal segment containing sufficient cues for humans to infer elevated risk before incident onset.
    \item \textbf{Visual evidence annotation:} identifying interpretable risk indicators used as reference evidence for evaluating model reasoning grounding (RGA).
\end{enumerate}

We conducted separate validation steps for guideline development, scalability audit, and crowdsroucing quality control. The agreement statistics reported in the main paper should not be interpreted as coming from a single annotation round.

\subsubsection{Annotation Protocol Development and Expert Consensus}

Before large-scale annotation, three internal researchers with experience in video understanding and safety-related visual analysis developed the annotation guidelines for crowdsourced data collection. The guidelines specified criteria for assigning risk/no-risk labels, principles for identifying early risk signal clips, and requirements for describing observable visual risk indicators.

The researchers independently annotated an initial set of 60 videos and subsequently discussed disagreements to refine the annotation protocol. During disagreement resolution, the experts examined cases where annotators differed in their interpretation of risk cues, temporal boundaries, or the level of specificity required for visual indicators.

For example, in a car-crash scenario where a vehicle was gradually drifting toward another lane, annotators initially disagreed on whether earlier frames showing minor lane deviation constituted sufficient evidence for a risk signal or whether the risk signal should begin only when the vehicle approached the neighboring car. The team refined the guideline to define the risk signal as the earliest point at which a reasonable observer could identify an elevated likelihood of an incident, rather than the moment when the collision became imminent.

Another example concerns natural-language risk indicators. Researchers refined the guideline to avoid speculative descriptions. For instance, descriptions such as ``the person intends to attack'' were replaced with observable descriptions such as ``the person is holding an object and moving rapidly toward another individual.'' These discussions were incorporated into the final annotation guidelines and examples provided to subsequent annotators.

After iterative refinement, the expert team reached 96\% consensus on this initial set. The refined guidelines, examples, and quality-control criteria were then used for subsequent data collection.

The following example illustrates the final annotation guideline for the arson domain.

\begin{tcolorbox}[
    breakable,
    title={Arson Domain Annotation Guideline}
]

    \textbf{Task Instruction}

    Imagine you are a detective now, and your job is to look into the future and tell us when you see clues that something is about to happen.

    Think about it like this: If you’re playing a game and someone is about to high-five you, they don't just magically touch your hand, right? First, they look at you. Then, they raise their hand back. Finally, they swing it forward. Those first two steps are clues (or signals) that tell you: "A high-five is coming".

    In this task, you’ll watch videos that might be Risky (like a fire starting where it shouldn't) or Benign (which just means "Totally Normal/Safe," like a backyard BBQ).

    \textbf{The Two Main Parts in Videos:}

    The "Clues" (Signal): This is the "Uh-oh" moment. It’s the part of the video you use to predict what happens next, e.g., seeing someone pouring liquid from a gas can, fumbling with a lighter near a trash can, or throwing a lit object, etc.

    The "Fire" (Event): The actual moment the fire catches and starts to spread, meaning the consequence of the clue/signal, e.g., the building, trash, or object actually starts burning (not just the lighter flame, but the real fire).

    \textbf{1. When does the "Clue/Signal" start and end?}

    \textbf{The Start Time [hh:mm:ss]:} This is the very first second you see someone preparing.

    If it’s a Risk (Dangerous): it’s the first second you see someone acting suspiciously (e.g., masking up, splashing liquid).

    If it’s Benign (Safe): It’s the second you see someone doing something normal (e.g., someone bringing out a bag of charcoal, opening a grill).

    \textbf{The End Time [hh:mm:ss]:} This is the very last second of the clue/signal, right before the fire catches.

    \textbf{2. What evidence are you seeing?}

    You need to prove why you made your prediction. In this part, you will list the specific things (objects) or behaviors (actions) that caught your eye during the "Clue" stage.

    If it's a RISK (Dangerous Fire), list the things that show someone is planning to start a fire on purpose to hurt property, such as:

    "Red gas can," "Lighter flicking," "Spilling gasoline on the floor instead of a car tank," "Breaking a window to throw something in," "Looking around to see if anyone is watching."

    If it's BENIGN (Safe/Normal Fire), list the things that show the fire is under control and meant for something good, like cooking or a party, such as:

    "BBQ Tongs," "Birthday cake," "Fireplace log," "Using a long safety lighter for a grill," "Singing 'Happy Birthday' before lighting a candle," "Arranging logs neatly inside a stone fire pit."

    \textbf{3. Tell the Story of the Clues/Signals}

    Now, you need to write a short description of the clues you just found. Imagine you are explaining what happened to someone who can't see the video. To make the timing clear, you must use timing words, such as First, Then, and Afterwards.

    If it’s a RISK (Dangerous):

    "First, a person in a hoodie approached the dumpster. Then, they poured liquid from a bottle onto paper. Afterwards, they reached for a lighter and threw it into the dumpster."

    If it’s BENIGN (Safe):

    "First, a person in an apron opened a grill. Then, they placed charcoal inside. Afterwards, they used a long lighter to start the briquettes."

    \textbf{4. When does the "Fire/Event" start?}

    \textbf{Fire Start Frame [hh:mm:ss]:} This is the exact moment the thing they were planning to light actually catches fire.

    If it's Risk (Dangerous): The second the building/trash/car starts burning.

    If it’s Benign (Safe): The second the charcoal or the candle wick catches fire.

    \textbf{Fire End Frame [hh:mm:ss]:} Mark the second when the fire goes out or the person leaves; if the video just cuts to black while the fire is still burning, use the very last second of the video.

    \textbf{5. Summarize the Fire Event}

    Risk (Dangerous):

    "The wooden shed was intentionally set on fire."

    Benign (Safe):

    "The family started a controlled backyard barbecue."

    \textbf{6. Was it Arson?}

    \textbf{Risk Label:}

    YES: If you saw someone intentionally trying to destroy property, or injure others, or start an uncontrolled fire.

    NO: If it was a person lighting a fire safely, normal fire situation a backyard grill, a professional firefighter doing a controlled burn, or someone lighting a candle.

\end{tcolorbox}

\subsubsection{Scalability Audit and Annotation Validation}

To evaluate whether the expert-developed annotation protocol could be consistently applied by non-expert annotators, we conducted a held-out scalability audit.

Two undergraduate annotators (one majoring in Anthropology and one in Computer Science) received three hours of training using the finalized annotation guidelines and example cases developed during the expert consensus stage. They then independently annotated a held-out set of 20 videos from the expert-labeled set described in Section~9.2.

For categorical risk labels (yes/no risk), both annotators achieved 100\% agreement with expert annotations. For temporal annotations, the annotators achieved a mean absolute deviation (MAD) of fewer than 4 frames compared with expert temporal boundaries. These results indicate that the annotation guidelines were sufficiently clear and could be reliably applied by trained external annotators beyond the original research team.

\subsubsection{Crowdsourcing Quality Control}

After validating the scalability of the annotation protocol, we conducted large-scale annotation through Amazon Mechanical Turk.

To ensure annotation quality, all workers first completed a mandatory qualification round consisting of 50 videos sampled from the expert-labeled set. Workers were retained for the final annotation pool only if they achieved greater than 98\% agreement with expert consensus for categorical labels and greater than 90\% agreement for timestamp annotations and natural-language risk indicators.

Following this qualification procedure, seven annotators were selected for the final annotation process. This multi-stage pipeline combines expert-defined annotation standards with scalable crowdsourcing while maintaining annotation reliability.

\begin{figure}[!h]
    \centering
    \includegraphics[width=\linewidth]{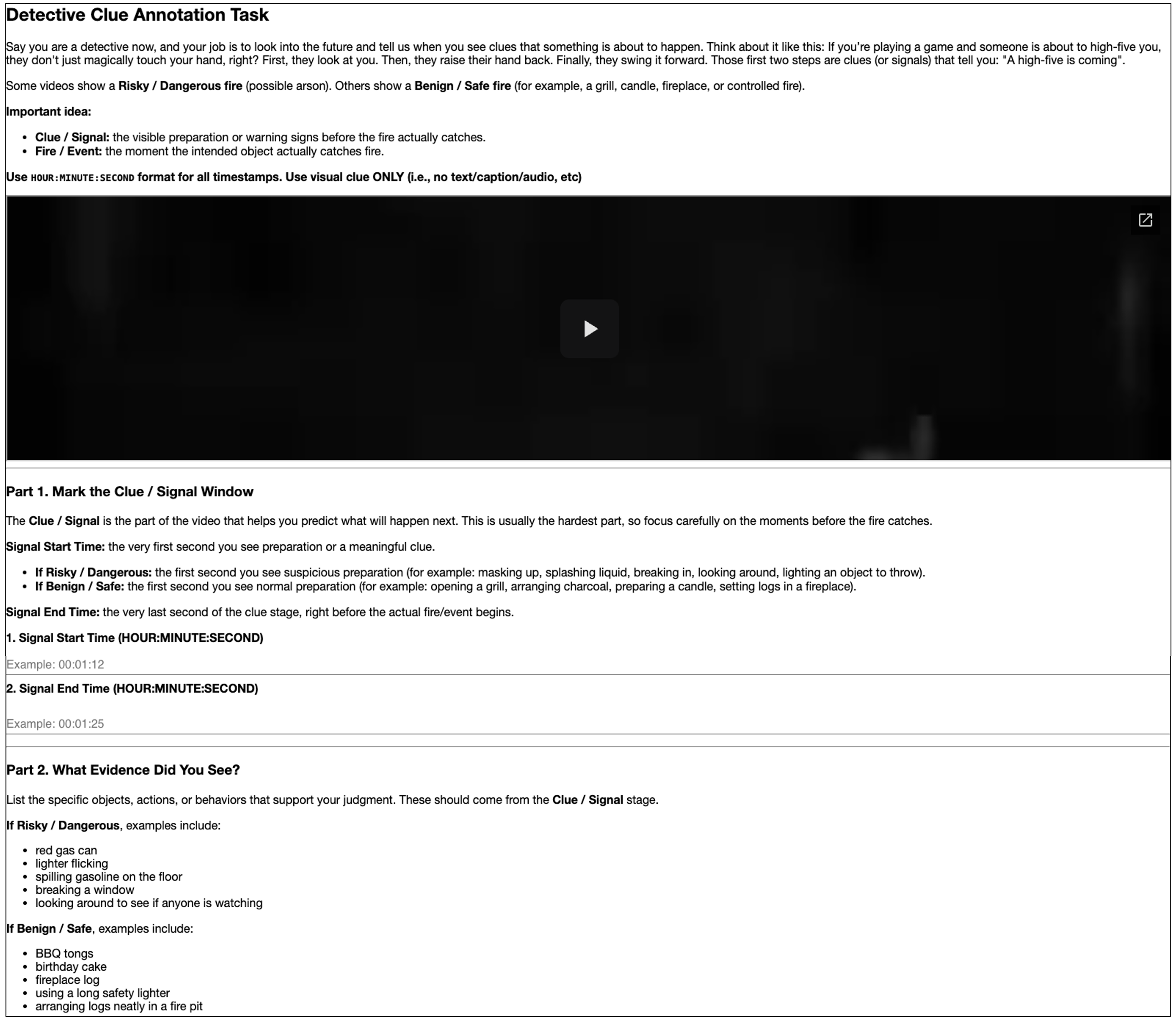}
    \caption{Crowdsourcing annotation interface for Arson. }
    \label{fig:annotation_interface}
\end{figure}

The crowdsourcing interface using Arson domain as an example is shown in Figure~\ref{fig:annotation_interface}. All annotators used the same standardized interface and annotation procedure.

\clearpage
\subsection{Additional Evaluation Metrics and F1 Analysis}
\label{app:additional_metrics}

The primary evaluation metric reported in the main paper is the positive-class F1 score, which captures the trade-off between precision and recall for identifying risky videos. However, as discussed by prior work, F1 alone does not fully capture classification performance, particularly under class imbalance or when true negatives are informative. Therefore, we provide additional evaluation metrics, including accuracy, specificity, balanced accuracy, Matthews correlation coefficient (MCC), and class-specific performance measures.

\subsubsection{Definition of Evaluation Metrics}

For each domain, we define the confusion matrix components as follows:

\begin{itemize}
    \item \textbf{True Positive (TP):} a risky video correctly predicted as risky.
    \item \textbf{False Positive (FP):} a non-risk video incorrectly predicted as risky.
    \item \textbf{True Negative (TN):} a non-risk video correctly predicted as non-risk.
    \item \textbf{False Negative (FN):} a risky video incorrectly predicted as non-risk.
\end{itemize}

The positive-class F1 score is computed as:

\begin{equation}
F1 = \frac{2 \times Precision \times Recall}
{Precision + Recall},
\end{equation}

where

\begin{equation}
Precision = \frac{TP}{TP+FP},
\qquad
Recall = \frac{TP}{TP+FN}.
\end{equation}

To complement F1, we additionally report accuracy, specificity, and balanced accuracy:

\begin{equation}
Accuracy =
\frac{TP+TN}{TP+TN+FP+FN}.
\end{equation}

\begin{equation}
Specificity =
\frac{TN}{TN+FP}.
\end{equation}

\begin{equation}
Balanced\ Accuracy =
\frac{Recall+Specificity}{2}.
\end{equation}

We also report Matthews correlation coefficient (MCC):

\begin{equation}
\small
MCC =
\frac{TP \times TN - FP \times FN}
{\sqrt{(TP+FP)(TP+FN)(TN+FP)(TN+FN)}} .
\end{equation}

MCC provides a balanced measure of binary classification performance by incorporating all four components of the confusion matrix and remains informative under class imbalance.

\subsubsection{Domain-Level Results}

Because RiskCueBench contains heterogeneous risk domains with different visual characteristics and class distributions, we report metrics separately for each domain rather than aggregating across domains into a single overall score. This avoids potential aggregation biases and preserves domain-specific performance differences.

Table~\ref{tab:additional_metrics} presents the additional evaluation metrics for all evaluated models.


\begin{table*}[t]
\centering
\caption{Additional evaluation metrics beyond positive-class F1 across RiskCueBench domains.}
\label{tab:additional_metrics}
\small
\begin{tabular}{llccccc}
\toprule
\textbf{Domain} &
\textbf{Model} &
\textbf{F1} &
\textbf{Accuracy} &
\textbf{Specificity} &
\textbf{Balanced Accuracy} &
\textbf{MCC} \\
\midrule

\multirow{11}{*}{Car Crash}
& Video-LLaVA      & 0.53 & 0.59 & 0.62 & 0.58 & 0.17 \\
& VideoLLaMA 3     & 0.59 & 0.59 & 0.60 & 0.59 & 0.18 \\
& Qwen3-VL-8B      & 0.64 & 0.61 & 0.61 & 0.61 & 0.22 \\
& MiMo-SFT         & 0.58 & 0.62 & 0.65 & 0.62 & 0.24 \\
& Apollo-7B        & 0.57 & 0.59 & 0.61 & 0.59 & 0.18 \\
& InternVL-3.5     & 0.61 & 0.64 & 0.66 & 0.64 & 0.28 \\
& Qwen3-VL-8B-T    & 0.66 & 0.65 & 0.62 & 0.65 & 0.29 \\
& MiMo-RL          & 0.63 & 0.64 & 0.66 & 0.64 & 0.28 \\
& Keye-VL 1.5      & 0.57 & 0.59 & 0.62 & 0.59 & 0.18 \\
& GLM 4.1 V-T      & 0.58 & 0.59 & 0.59 & 0.59 & 0.18 \\
& Gemini-Flash-3   & 0.69 & 0.64 & 0.58 & 0.64 & 0.27 \\
\midrule

\multirow{11}{*}{Protest}
& Video-LLaVA      & 0.44 & 0.64 & 0.70 & 0.60 & 0.18 \\
& VideoLLaMA 3     & 0.46 & 0.67 & 0.70 & 0.63 & 0.24 \\
& Qwen3-VL-8B      & 0.48 & 0.62 & 0.65 & 0.60 & 0.20 \\
& MiMo-SFT         & 0.47 & 0.61 & 0.65 & 0.59 & 0.17 \\
& Apollo-7B        & 0.45 & 0.60 & 0.65 & 0.58 & 0.15 \\
& InternVL-3.5     & 0.47 & 0.63 & 0.67 & 0.60 & 0.20 \\
& Qwen3-VL-8B-T    & 0.57 & 0.62 & 0.66 & 0.62 & 0.24 \\
& MiMo-RL          & 0.51 & 0.60 & 0.65 & 0.59 & 0.18 \\
& Keye-VL 1.5      & 0.59 & 0.59 & 0.58 & 0.59 & 0.17 \\
& GLM 4.1 V-T      & 0.56 & 0.59 & 0.61 & 0.59 & 0.18 \\
& Gemini-Flash-3   & 0.67 & 0.65 & 0.67 & 0.66 & 0.31 \\
\midrule

\multirow{11}{*}{Arson}
& Video-LLaVA      & 0.55 & 0.59 & 0.61 & 0.59 & 0.18 \\
& VideoLLaMA 3     & 0.57 & 0.61 & 0.62 & 0.60 & 0.21 \\
& Qwen3-VL-8B      & 0.62 & 0.61 & 0.61 & 0.61 & 0.21 \\
& MiMo-SFT         & 0.56 & 0.57 & 0.58 & 0.57 & 0.13 \\
& Apollo-7B        & 0.58 & 0.64 & 0.67 & 0.63 & 0.26 \\
& InternVL-3.5     & 0.60 & 0.64 & 0.65 & 0.64 & 0.28 \\
& Qwen3-VL-8B-T    & 0.64 & 0.62 & 0.60 & 0.62 & 0.24 \\
& MiMo-RL          & 0.61 & 0.62 & 0.61 & 0.62 & 0.23 \\
& Keye-VL 1.5      & 0.59 & 0.63 & 0.64 & 0.62 & 0.25 \\
& GLM 4.1 V-T      & 0.58 & 0.60 & 0.59 & 0.60 & 0.19 \\
& Gemini-Flash-3   & 0.62 & 0.61 & 0.59 & 0.61 & 0.21 \\
\midrule

\multirow{11}{*}{Cruise Collision}
& Video-LLaVA      & 0.55 & 0.57 & 0.60 & 0.57 & 0.15 \\
& VideoLLaMA 3     & 0.56 & 0.59 & 0.60 & 0.59 & 0.18 \\
& Qwen3-VL-8B      & 0.63 & 0.61 & 0.62 & 0.61 & 0.22 \\
& MiMo-SFT         & 0.59 & 0.59 & 0.59 & 0.59 & 0.19 \\
& Apollo-7B        & 0.57 & 0.58 & 0.61 & 0.58 & 0.16 \\
& InternVL-3.5     & 0.59 & 0.63 & 0.66 & 0.63 & 0.26 \\
& Qwen3-VL-8B-T    & 0.66 & 0.63 & 0.62 & 0.63 & 0.26 \\
& MiMo-RL          & 0.64 & 0.63 & 0.64 & 0.63 & 0.26 \\
& Keye-VL 1.5      & 0.57 & 0.58 & 0.60 & 0.58 & 0.16 \\
& GLM 4.1 V-T      & 0.56 & 0.59 & 0.59 & 0.59 & 0.18 \\
& Gemini-Flash-3   & 0.68 & 0.67 & 0.63 & 0.66 & 0.33 \\

\bottomrule
\end{tabular}
\end{table*}

\subsubsection{Comparison with No-Skill Baselines}

To contextualize positive-class F1 values, we additionally compare model performance against an always-risk classifier, which predicts all samples as risky. This classifier provides a no-skill baseline that depends on the prevalence of positive examples in each domain.

\begin{table}[h!]
    \centering
    \small
    \begin{tabular}{lcc}
    \toprule
    \textbf{Domain} & \textbf{Always-Risk F1} & \textbf{Random Guess F1} \\
    \midrule
    Car Crash          & 0.668 & 0.50 \\
    Protest             & 0.619 & 0.44 \\
    Arson               & 0.722 & 0.53 \\
    Cruise Collision    & 0.640 & 0.47 \\
    \bottomrule
    \end{tabular}
    \caption{No-skill baseline comparison across RiskCueBench domains. The always-risk classifier predicts all samples as risky, providing a prevalence-dependent positive-class F1 baseline.}
    \label{tab:no_skill_baseline}
\end{table}

The additional no-skill F1 strengthens rather than changes this central conclusion as in \ref{tab:no_skill_baseline}. Several evaluated models do not exceed these values. This shows that positive-class F1 alone may provide an overly favorable impression of model performance because it rewards high recall without accounting for true-negative performance.

\subsubsection{Robustness of Model Ranking}

To examine whether conclusions depend on the choice of evaluation metric, we compare model rankings across positive-class F1, balanced accuracy, and MCC.


The relative ordering of models remains consistent across these complementary metrics, suggesting that the observed performance gap between current VLMs and human performance is not an artifact of the choice of F1 alone.

\clearpage
\subsection{Length-Matched Analysis}
\label{app:length_matching}

To isolate the effect of clip duration, we constructed full-length-matched positive (risk) and negative (non-risk) video subsets within each domain and repeated the evaluation using the same protocol as the main experiments. Model performance remained highly stable after duration matching.

Across all evaluated models, the absolute change in F1 was no greater than 0.01 in either domain. RGA, TRD, and SCD also showed only minor changes relative to the full-sample results. For example, Car Crash F1 ranged from 0.52 to 0.65 in the matched analysis, compared with 0.53 to 0.66 in the original analysis, while Protest F1 ranged from 0.43 to 0.58, compared with 0.44 to 0.59 originally.

These results suggest that the observed performance limitations are not primarily explained by differences in clip duration between risk and no-risk videos. We provide the complete length-matched results below.

\begin{table*}[t]
\centering
\small
\resizebox{\textwidth}{!}{
\begin{tabular}{llcccccccc}
\toprule
\textbf{Model Category} & \textbf{Model} 
& \multicolumn{4}{c}{\textbf{Car Crash}} 
& \multicolumn{4}{c}{\textbf{Protest}} \\
\cmidrule(lr){3-6}
\cmidrule(lr){7-10}
& 
& \textbf{F1} & \textbf{RGA} & \textbf{TRD} & \textbf{SCD}
& \textbf{F1} & \textbf{RGA} & \textbf{TRD} & \textbf{SCD}\\
\midrule

\multirow{5}{*}{Non-Reasoning}
& Video-LLaVA    
& 0.52 & 47.6 & 1.90 & -- 
& 0.43 & 32.9 & 1.92 & -- \\
& VideoLLaMA 3   
& 0.58 & 53.4 & 1.69 & -- 
& 0.45 & 38.7 & 2.49 & -- \\
& Qwen3-VL-8B    
& 0.63 & 56.8 & 1.46 & -- 
& 0.47 & 42.2 & 2.28 & -- \\
& MiMo-SFT       
& 0.57 & 50.6 & 2.17 & -- 
& 0.48 & 37.4 & 1.81 & -- \\
& Apollo-7B      
& 0.58 & 52.9 & 1.89 & -- 
& 0.44 & 36.8 & 2.66 & -- \\

\midrule

\multirow{5}{*}{Reasoning}
& InternVL-3.5     
& 0.60 & 63.4 & 2.94 & -0.27 
& 0.46 & 47.1 & 2.61 & -0.41 \\
& Qwen3-VL-8B-T    
& 0.65 & 68.0 & 2.26 & -0.28 
& 0.56 & 55.8 & 2.14 & -0.35 \\
& MiMo-RL          
& 0.62 & 60.7 & 3.75 & -0.36 
& 0.50 & 50.1 & 3.50 & -0.44 \\
& Keye-VL 1.5      
& 0.56 & 64.0 & 3.20 & -0.20 
& 0.58 & 51.6 & 2.97 & -0.30 \\
& GLM 4.1 V-T      
& 0.57 & 64.4 & 3.35 & -0.25 
& 0.55 & 48.4 & 3.25 & -0.33 \\

\bottomrule
\end{tabular}
}
\caption{Length-matched evaluation results for Car Crash and Protest domains. After matching risk and non-risk videos by duration, model performance remains comparable to the original evaluation, suggesting that observed performance differences are not primarily driven by duration imbalance.}
\label{tab:length_matched}
\end{table*}

\clearpage
\subsection{Overlap of Qualitative Error Categories}
\label{app:error_overlap}

Since the qualitative analysis was conducted exclusively on misclassified cases, conclusion errors occur in all 336 samples by construction. The overlap analysis therefore primarily characterizes the relationship between perceptual and reasoning failures preceding the incorrect conclusion.

\begin{table}[h]
\centering
\small
\begin{tabular}{lccc}
\toprule
 & \textbf{Perceptual} & \textbf{Reasoning} & \textbf{Conclusion} \\
\midrule
\textbf{Perceptual}
& 192 (57.1\%)
& 31 (9.2\%)
& 192 (57.1\%) \\

\textbf{Reasoning}
& 31 (9.2\%)
& 65 (19.3\%)
& 65 (19.3\%) \\

\textbf{Conclusion}
& 192 (57.1\%)
& 65 (19.3\%)
& 336 (100\%) \\
\bottomrule
\end{tabular}
\caption{Overlap matrix for qualitative error categories among 336 misclassified cases. Diagonal entries report the prevalence of each error category, while off-diagonal entries report pairwise co-occurrence. Because all analyzed cases are prediction failures, conclusion errors occur in every sample.}
\label{tab:error_overlap}
\end{table}

\clearpage
\subsection{Key-Frame Analysis for Temporal vs. Static Cue Reliance}
\label{app:keyframe_analysis}

To further distinguish temporal-order sensitivity from reliance on sparse static cues, we conducted an extreme-sparse key-frame analysis. Specifically, we evaluated models under two additional conditions: (1) First Frame Only, where models received only the earliest frame of the risk signal clip, and (2) Last Frame Only, where models received only the final frame immediately preceding the incident. These conditions provide a comparison between early static evidence, late-stage static evidence, and the full temporal information available in the original video condition.

Across domains, the first-frame condition consistently underperformed the full-video condition, with an average F1 decrease of 0.015--0.033. This suggests that the earliest static frame generally does not contain sufficient information for reliable risk prediction. In contrast, last-frame performance was generally comparable to or slightly higher than full-video performance. This pattern is expected because the last frame represents a privileged condition immediately before incident onset, where the most explicit pre-incident visual cues are likely to appear.

These results clarify the interpretation of the Temporal Reasoning Difference (TRD) metric. TRD is intended to measure sensitivity to temporal ordering rather than whether static visual cues contain any predictive information. Since later frames may already contain highly salient risk indicators, strong performance from the last-frame condition does not imply effective temporal reasoning. Instead, the combination of limited performance differences under temporal perturbations and comparable last-frame performance suggests that current VLMs rely substantially on salient visual states while making limited use of the temporal progression through which these states emerge.

\begin{table*}[t]
\centering
\scriptsize
\setlength{\tabcolsep}{2.5pt}
\renewcommand{\arraystretch}{1.05}

\resizebox{\textwidth}{!}{%
\begin{tabular}{lcccccccccccc}
\toprule
\textbf{Model}
& \textbf{CC V} & \textbf{CC 1st} & \textbf{CC Last}
& \textbf{PR V} & \textbf{PR 1st} & \textbf{PR Last}
& \textbf{AR V} & \textbf{AR 1st} & \textbf{AR Last}
& \textbf{CR V} & \textbf{CR 1st} & \textbf{CR Last} \\
\midrule

Video-LLaVA
& .53 & .49 & .55
& .44 & .42 & .43
& .55 & .50 & .58
& .55 & .50 & .58 \\

VideoLLaMA 3
& .59 & .55 & .61
& .46 & .44 & .45
& .57 & .52 & .60
& .56 & .51 & .59 \\

Qwen3-VL-8B
& .64 & .61 & .63
& .48 & .47 & .46
& .62 & .59 & .61
& .63 & .60 & .62 \\

MiMo-SFT
& .58 & .52 & .62
& .47 & .43 & .49
& .56 & .50 & .60
& .59 & .53 & .63 \\

Apollo-7B
& .57 & .53 & .59
& .45 & .43 & .44
& .58 & .53 & .61
& .57 & .52 & .60 \\

InternVL-3.5
& .61 & .63 & .60
& .47 & .49 & .46
& .60 & .61 & .59
& .59 & .60 & .58 \\

Qwen3-VL-8B-T
& .66 & .66 & .66
& .57 & .57 & .57
& .64 & .63 & .64
& .66 & .65 & .66 \\

MiMo-RL
& .63 & .58 & .67
& .51 & .47 & .54
& .61 & .56 & .65
& .64 & .59 & .68 \\

Keye-VL 1.5
& .57 & .56 & .58
& .59 & .58 & .59
& .59 & .57 & .60
& .57 & .55 & .58 \\

GLM 4.1 V-T
& .58 & .57 & .59
& .56 & .55 & .56
& .58 & .56 & .59
& .56 & .54 & .57 \\

\midrule
\textbf{Mean}
& .596 & .570 & .610
& .500 & .485 & .499
& .590 & .557 & .607
& .592 & .559 & .609 \\

\bottomrule
\end{tabular}%
}

\caption{
Key-frame analysis across RiskCueBench domains.
CC, PR, AR, and CR denote Car Crash, Protest, Arson, and Cruise Collision, respectively.
V denotes the original full-video condition, while 1st and Last denote extreme-sparse baselines using only the first or last frame of the risk signal clip.
}
\label{tab:keyframe_analysis}
\end{table*}

\clearpage
\subsection{Human Validation of LLM-Based Reasoning Grounding Evaluation}
\label{app:rga_validation}

The Reasoning Grounding Accuracy (RGA) metric evaluates whether the visual evidence mentioned in a model-generated reasoning chain corresponds to human-annotated risk indicators. Because manually evaluating reasoning grounding at the scale of all model outputs is costly, we use an LLM-based evaluator to extract decision items from model reasoning and compute their semantic alignment with annotated visual indicators. To validate the reliability of this approach, we conducted a human audit comparing samples with high and low RGA scores.

\subsubsection{LLM-Based RGA Evaluation}

For each model-generated reasoning chain, the LLM evaluator extracts concrete visual decision items, including objects, actions, and observable cues referenced by the model as evidence for its prediction. These extracted items are compared with human-annotated visual risk indicators using embedding-based semantic similarity. The resulting alignment score is aggregated to produce the RGA metric.

The LLM evaluator enables large-scale analysis across all model outputs while maintaining a consistent evaluation protocol. However, recognizing that LLM-based evaluation may introduce noise, we performed an independent human validation analysis.

\subsubsection{Human Validation Protocol}

We sampled reasoning outputs from the Protest domain, which contains socially embedded risk scenarios and represents one of the more challenging domains in RiskCueBench. We selected 70 samples consisting of 35 samples above the 75th percentile and 35 samples below the 25th percentile.

For each sampled reasoning output, human annotators examined whether the visual evidence items extracted by the LLM evaluator semantically corresponded to verified visual indicators present in the video and human annotations. Annotators were instructed to judge whether the referenced evidence represented observable visual cues rather than speculative interpretations or unsupported assumptions.

\subsubsection{Validation Results}

Human evaluation showed a clear separation between high- and low-RGA samples. Among reasoning items classified as grounded by RGA, 75\% were confirmed by human annotators to correspond to verified visual risk indicators. In contrast, only 21\% of reasoning items classified as ungrounded by RGA contained evidence that humans considered visually supported.

These results indicate that RGA provides a meaningful approximation of visual grounding quality. While LLM-based evaluation may introduce occasional errors, the human validation demonstrates that RGA successfully distinguishes reasoning chains that reference observable visual evidence from those that rely on unsupported or irrelevant explanations.

\begin{table}[t]
\centering
\small
\setlength{\tabcolsep}{5pt}
\begin{tabular}{lcc}
\toprule
\textbf{RGA Group} & \textbf{Verified (\%)} & \textbf{$N$} \\
\midrule
High RGA & 75 & 35 \\
Low RGA  & 21 & 35 \\
\bottomrule
\end{tabular}
\caption{Human validation of RGA. Verified indicates the percentage of reasoning items judged by human annotators to correspond to visual evidence in the video and annotations.}
\label{tab:rga_human_validation}
\end{table}

\end{document}